\documentclass{l4dc2027}
\usepackage{times}
\usepackage{booktabs}
\usepackage{tikz}
\usetikzlibrary{arrows.meta,calc,positioning}
\usepackage{pifont}

\definecolor{keyc}{HTML}{1D5FC4}
\definecolor{okc}{HTML}{1A7A43}
\definecolor{badc}{HTML}{B3302A}
\definecolor{tbdc}{HTML}{C05600}

\newtheorem{assumption}{Assumption}
\expandafter\def\csname res@act/W3/NoAct\endcsname{217}
\expandafter\def\csname res@actEps\endcsname{0.2}
\expandafter\def\csname res@ax/W1/Flat/air\endcsname{76.2}
\expandafter\def\csname res@ax/W1/Onto/air\endcsname{1.9}
\expandafter\def\csname res@cf/W1/Naive\endcsname{266}
\expandafter\def\csname res@cfg/T\endcsname{400}
\expandafter\def\csname res@cfg/clip\endcsname{0.2}
\expandafter\def\csname res@cfg/ent\endcsname{0.001}
\expandafter\def\csname res@cfg/epochs\endcsname{5}
\expandafter\def\csname res@cfg/eps_per_iter\endcsname{40}
\expandafter\def\csname res@cfg/gamma\endcsname{0.99}
\expandafter\def\csname res@cfg/iters\endcsname{200}
\expandafter\def\csname res@cfg/lam\endcsname{0.95}
\expandafter\def\csname res@cfg/lr\endcsname{0.001}
\expandafter\def\csname res@cfg/minibatch\endcsname{16384}
\expandafter\def\csname res@ddelMax\endcsname{$+0.04$}
\expandafter\def\csname res@ddelMin\endcsname{$-0.11$}
\expandafter\def\csname res@depMedHi\endcsname{17}
\expandafter\def\csname res@depMedLo\endcsname{10}
\expandafter\def\csname res@dvs/W3\endcsname{$-6.98$ [$-12.47$, $-1.70$]}
\expandafter\def\csname res@dvsMax\endcsname{$-2.3$}
\expandafter\def\csname res@dvsMin\endcsname{$-8.3$}
\expandafter\def\csname res@effFlat\endcsname{18}
\expandafter\def\csname res@effFlatW3\endcsname{5}
\expandafter\def\csname res@effOnto\endcsname{10}
\expandafter\def\csname res@effOntoW3\endcsname{3}
\expandafter\def\csname res@fcDelW1\endcsname{65}
\expandafter\def\csname res@fcDelW3\endcsname{91}
\expandafter\def\csname res@flatAlpha\endcsname{1.4}
\expandafter\def\csname res@flatHumanD\endcsname{0.5}
\expandafter\def\csname res@flatPayD\endcsname{0.8}
\expandafter\def\csname res@gammaMain\endcsname{1}
\expandafter\def\csname res@gn/1/p\endcsname{0.63}
\expandafter\def\csname res@gn/1/vpN\endcsname{23}
\expandafter\def\csname res@gn/10/p\endcsname{$<$0.001}
\expandafter\def\csname res@gn/10/vpN\endcsname{44}
\expandafter\def\csname res@gn/10/vpO\endcsname{21}
\expandafter\def\csname res@gn/3/vpN\endcsname{27}
\expandafter\def\csname res@gnOntoHi\endcsname{21}
\expandafter\def\csname res@gnOntoLo\endcsname{20}
\expandafter\def\csname res@gselDelHi\endcsname{2.02}
\expandafter\def\csname res@gselDelLo\endcsname{1.76}
\expandafter\def\csname res@gselGammas\endcsname{1, 1.5, 2, 3, 5}
\expandafter\def\csname res@gselList\endcsname{14, 15, 17, 17, 24}
\expandafter\def\csname res@incDelivOnto\endcsname{9.90}
\expandafter\def\csname res@incDelivRef\endcsname{10.68}
\expandafter\def\csname res@incDet\endcsname{100}
\expandafter\def\csname res@incHermit\endcsname{10/10}
\expandafter\def\csname res@incN\endcsname{100}
\expandafter\def\csname res@incOntoEps\endcsname{4}
\expandafter\def\csname res@incOntoSteps\endcsname{78}
\expandafter\def\csname res@incTrustCf\endcsname{5}
\expandafter\def\csname res@incTrustEps\endcsname{60}
\expandafter\def\csname res@incTrustSteps\endcsname{1995}
\expandafter\def\csname res@infRedMax\endcsname{41}
\expandafter\def\csname res@infRedMin\endcsname{18}
\expandafter\def\csname res@jvDesignPct\endcsname{94.4}
\expandafter\def\csname res@jvHeavyFail\endcsname{853}
\expandafter\def\csname res@jvOwnPct\endcsname{91.6}
\expandafter\def\csname res@jvSplitPct\endcsname{92.4}
\expandafter\def\csname res@jvStates\endcsname{78{,}837}
\expandafter\def\csname res@margin\endcsname{0.05}
\expandafter\def\csname res@mu\endcsname{1000}
\expandafter\def\csname res@nBoot\endcsname{10{,}000}
\expandafter\def\csname res@nRunCells\endcsname{12}
\expandafter\def\csname res@nRuns\endcsname{3}
\expandafter\def\csname res@nSeeds\endcsname{100}
\expandafter\def\csname res@nViolDownRuns\endcsname{4}
\expandafter\def\csname res@notCloseHi\endcsname{68}
\expandafter\def\csname res@notCloseLo\endcsname{56}
\expandafter\def\csname res@rhoNeed\endcsname{6.34}
\expandafter\def\csname res@rhoNeedCls\endcsname{HeavyDrone}
\expandafter\def\csname res@rw/arrive\endcsname{1}
\expandafter\def\csname res@rw/infeas\endcsname{0.5}
\expandafter\def\csname res@rw/interv\endcsname{0.01}
\expandafter\def\csname res@rw/progress\endcsname{1}
\expandafter\def\csname res@rw/speed\endcsname{0.5}
\expandafter\def\csname res@rw/viol\endcsname{0.2}
\expandafter\def\csname res@singleMinVp\endcsname{79}
\expandafter\def\csname res@stCfFirstPct\endcsname{13}
\expandafter\def\csname res@stGcfFirst\endcsname{3}
\expandafter\def\csname res@stGfirst\endcsname{10}
\expandafter\def\csname res@stGmax\endcsname{30}
\expandafter\def\csname res@stN\endcsname{91}
\expandafter\def\csname res@stNaiveCf\endcsname{100}
\expandafter\def\csname res@stNaiveCfMain\endcsname{1}
\expandafter\def\csname res@stNaiveDmin\endcsname{0.75}
\expandafter\def\csname res@stNaiveViol\endcsname{48}
\expandafter\def\csname res@stOntoDmin\endcsname{1.12}
\expandafter\def\csname res@tCompiled\endcsname{0.06}
\expandafter\def\csname res@tHermit\endcsname{332}
\expandafter\def\csname res@testTrials\endcsname{72}
\expandafter\def\csname res@tfMax\endcsname{6.1}
\expandafter\def\csname res@trMax\endcsname{4.6}
\expandafter\def\csname res@untypedOrdPct\endcsname{66}
\expandafter\def\csname res@vcBaseHi\endcsname{3.3}
\expandafter\def\csname res@vcBaseLo\endcsname{1.9}
\expandafter\def\csname res@vcJointShare\endcsname{98}
\expandafter\def\csname res@vp/W0/Flat\endcsname{20}
\expandafter\def\csname res@vp/W0/Onto\endcsname{22}
\expandafter\def\csname res@vp/W3/FlatCons\endcsname{28}
\expandafter\def\csname res@vp/W3/Onto\endcsname{40}
\expandafter\def\csname res@vpFlatUnseenMin\endcsname{89}
\expandafter\def\csname res@vpOntoMax\endcsname{70}
\expandafter\def\csname res@vpOntoMin\endcsname{20}
\expandafter\def\csname res@vpRLMax\endcsname{65}
\expandafter\def\csname res@vpRLMin\endcsname{11}
\expandafter\def\csname res@vsNoneMax\endcsname{754}
\expandafter\def\csname res@vsNoneMin\endcsname{551}
\expandafter\def\csname res@vsOntoMax\endcsname{47.9}
\expandafter\def\csname res@vsOntoMin\endcsname{5.6}
\expandafter\def\csname res@vsrcOnsetsDistinct\endcsname{466}

\newcommand{\res}[1]{\ifcsname res@#1\endcsname\csname res@#1\endcsname\else\textcolor{red}{\textbf{??}}\fi}

\newcommand{\cls}[1]{\ensuremath{\mathsf{#1}}}
\newcommand{\T}{\mathcal{T}}
\newcommand{\A}{\mathcal{A}}
\newcommand{\K}{\mathcal{K}}
\newcommand{\X}{\mathcal{X}}
\newcommand{\dyn}{\preceq_{\mathrm{dyn}}}
\newcommand{\ndyn}{\not\preceq_{\mathrm{dyn}}}
\newcommand{\cmark}{\ding{51}}
\newcommand{\xmark}{\ding{55}}
\newcommand{\pmark}{$\circ$}

\usepackage{rotating}
\makeatletter
\renewcommand\ps@jmlrtps{\let\@mkboth\@gobbletwo\def\@oddhead{\scriptsize Preprint.\hfill}\let\@evenhead\@oddhead\let\@oddfoot\@empty\let\@evenfoot\@empty}
\renewcommand*{\@jmlrenddoc}{\FloatBarrier}
\makeatother

\title[Ontology-Compiled Control Barrier Functions]{Ontology-Compiled Control Barrier Functions for Heterogeneous Robot Teams}

\author[Wu et al.]{%
 \Name{Peng Wu\nametag{$^{1}$}} \Email{p.wu@northeastern.edu}\\
 \Name{Seyede Fatemeh Ghoreishi\nametag{$^{1}$}}, \Name{Nathan Bastian\nametag{$^{2}$}}, \Name{Mohsen Imani\nametag{$^{3}$}}, \Name{Tian Lan\nametag{$^{4}$}}, \Name{Mahdi Imani\nametag{$^{1}$}}\\
 \addr $^{1}$Northeastern University, Boston, MA, USA \quad $^{2}$Syracuse University, Syracuse, NY, USA\\
 \addr $^{3}$University of California, Irvine, CA, USA \quad $^{4}$The George Washington University, Washington, DC, USA%
}

\begin{document}

\maketitle

\begin{abstract}%
Safety filters for heterogeneous robot teams fail because of missing constraints, overestimated braking capabilities and conflicts between individually satisfiable constraints.
We present OntoCBF, which compiles a domain ontology with a class taxonomy, typed relations, safety axioms and a dynamics profile per robot class into a control barrier function filter with one quadratic program per agent.
Description-logic entailment supplies the constraints of any team online, including those that follow from chains of relations and from classes defined by what a robot carries, so the existing safety axioms cover new robot classes, payload classes and combinations.
We prove that a subclass may reuse the braking capability of its parent whenever it can produce every velocity of the parent, a property separate from subsumption.
For box-bounded double integrators this property is also necessary, and every loss of safety then begins with a conflict.
Once three robots interact, the safe set of pairwise barriers contains states from which a collision is unavoidable, and such states persist in larger teams under any fixed capability reduction.
In a simulated warehouse where humans ignore the robots, a filter written for the design classes violates safety in \res{vpFlatUnseenMin}--100\,\% of the episodes with unseen classes.
In all scenarios, including one with 32 robots, OntoCBF violates safety in \res{vpOntoMin}--\res{vpOntoMax}\,\% of the episodes, each time after a conflict, with a median deepest intrusion of \res{depMedLo}--\res{depMedHi}\,\% of the keep-out distance.
A multi-agent PPO policy trained with the filter in the loop lowers the rate of infeasible quadratic programs, nearly all of them conflicts, by \res{infRedMin}--\res{infRedMax}\,\%.
\end{abstract}


\section{Introduction}\label{sec:intro}

Warehouses, hospitals and construction sites increasingly operate heterogeneous robot teams in spaces shared with people.
Their safety requirements depend on what each entity is and on how the entities are related.
A drone keeps farther from a person than a ground robot does, a carrier is responsible for the distance between its payload and nearby people, and ground robots avoid restricted zones that drones may cross.
Industrial standards state such requirements in the same terms, through separation distances that depend on the type of machine and on its stopping performance \citep{iso15066,iso36914}.
Safety filters based on control barrier functions (CBFs) enforce requirements of this kind on top of a learned or hand-designed controller with minimal changes to its commands \citep{ames2017cbf,wang2017barrier}, and they are a common safety layer for multi-agent reinforcement learning (MARL) \citep{cai2021decentralized,zhang2025gcbfplus}.

In existing multi-agent safety filters, the constrained pairs, the keep-out distances and the braking capabilities are fixed at design time for the robot types known at that point \citep{wang2017barrier,cai2021decentralized,elsayedaly2021shield}.
Learned graph CBFs for collision avoidance generalize across team sizes and, empirically, across robot models \citep{zhang2025gcbfplus,kim2026cchgcbf}.
Semantic safety filters query language models for constraints, which leaves the completeness and consistency of the constraint set unverified \citep{brunke2025semantic,ravichandran2026core}.
Robot fleets change over their lifetime, and whenever a team gains a new robot model, payload type or task combination, an engineer has to determine by hand which constraints apply to the new entities and whether the existing safety certificate still holds.

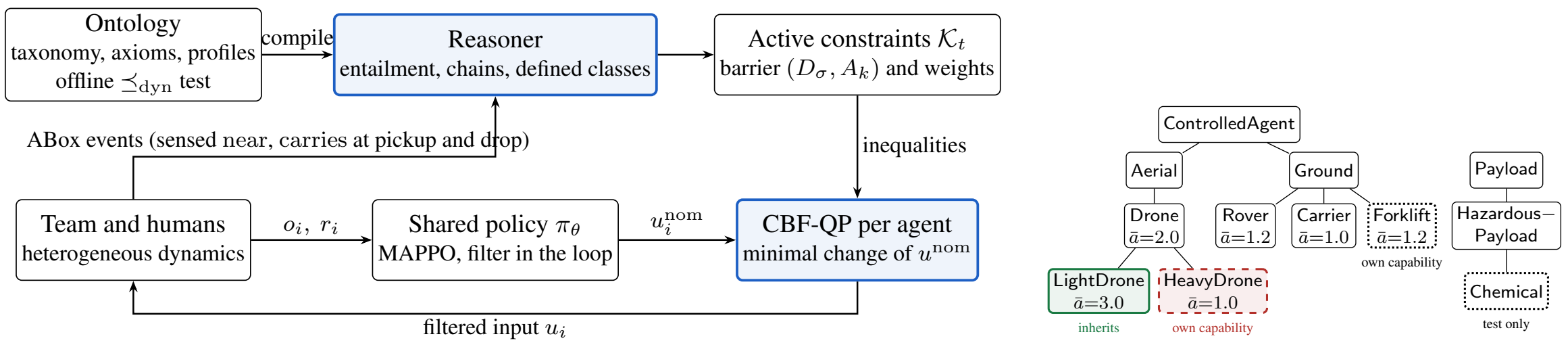
\begin{figure}[t]
\centering
\begin{minipage}[b]{0.64\linewidth}\centering
\resizebox{\linewidth}{!}{%
\begin{tikzpicture}[font=\footnotesize,
  box/.style={draw, rounded corners=2pt, align=center, minimum width=2.6cm, minimum height=1.0cm, inner sep=2pt},
  key/.style={box, draw=keyc, line width=0.9pt, fill=keyc!7},
  arr/.style={-{Stealth[length=4pt]}, semithick},
  lab/.style={font=\scriptsize, inner sep=1.5pt}]
  \node[box] (ont) at (0,2.3) {Ontology\\[-1pt]{\scriptsize taxonomy, axioms, profiles}\\[-1pt]{\scriptsize offline $\dyn$ test}};
  \node[key] (rea) at (4.5,2.3) {Reasoner\\[-1pt]{\scriptsize entailment, chains, defined classes}};
  \node[box] (con) at (9.0,2.3) {Active constraints $\K_t$\\[-1pt]{\scriptsize barrier $(D_\sigma,A_k)$ and weights}};
  \node[box] (env) at (0,0) {Team and humans\\[-1pt]{\scriptsize heterogeneous dynamics}};
  \node[box] (pol) at (4.5,0) {Shared policy $\pi_\theta$\\[-1pt]{\scriptsize MAPPO, filter in the loop}};
  \node[key] (qp) at (9.0,0) {CBF-QP per agent\\[-1pt]{\scriptsize minimal change of $u^{\mathrm{nom}}$}};
  \draw[arr] (ont) -- node[lab,above]{compile} (rea);
  \draw[arr] (rea) -- (con);
  \draw[arr] (con) -- node[lab,right]{inequalities} (qp);
  \draw[arr] (env) -- node[lab,above]{$o_i,\ r_i$} (pol);
  \draw[arr] (pol) -- node[lab,above]{$u_i^{\mathrm{nom}}$} (qp);
  \draw[arr] (qp.south) -- ++(0,-0.4) -| node[lab,pos=0.25,below]{filtered input $u_i$} (env.south);
  \draw[arr] (env.north) |- node[lab,pos=0.7,above]{ABox events (sensed $\mathrm{near}$, $\mathrm{carries}$ at pickup and drop)} ($(rea.south)+(0,-0.75)$) -- (rea.south);
\end{tikzpicture}}\\[2pt]
{\small (a) Pipeline}
\end{minipage}\hfill
\begin{minipage}[b]{0.33\linewidth}\centering
\resizebox{\linewidth}{!}{%
\begin{tikzpicture}[font=\scriptsize,
  c/.style={draw, rounded corners=2pt, inner sep=2pt, minimum height=0.55cm, align=center},
  ok/.style={c, draw=okc, line width=0.9pt, fill=okc!8},
  bad/.style={c, draw=badc, line width=0.9pt, fill=badc!8, dashed},
  new/.style={c, line width=0.9pt, densely dotted}]
  \node[c] (ca) at (2.0,2.55) {\cls{ControlledAgent}};
  \node[c] (ae) at (0.85,1.85) {\cls{Aerial}};
  \node[c] (gr) at (3.45,1.85) {\cls{Ground}};
  \node[c] (dr) at (0.85,1.0) {\cls{Drone}\\$\bar a{=}2.0$};
  \node[c] (ro) at (2.25,1.0) {\cls{Rover}\\$\bar a{=}1.2$};
  \node[c] (cr) at (3.45,1.0) {\cls{Carrier}\\$\bar a{=}1.0$};
  \node[ok] (li) at (0.0,0.0) {\cls{LightDrone}\\$\bar a{=}3.0$};
  \node[bad] (he) at (1.75,0.0) {\cls{HeavyDrone}\\$\bar a{=}1.0$};
  \node[new] (fo) at (4.65,1.0) {\cls{Forklift}\\$\bar a{=}1.2$};
  \draw (ca) -- (ae); \draw (ca) -- (gr);
  \draw (ae) -- (dr); \draw (gr) -- (ro); \draw (gr) -- (cr);
  \draw (dr) -- (li); \draw (dr) -- (he); \draw (gr) -- (fo);
  \node[font=\tiny, text=okc, below=0pt of li] {inherits};
  \node[font=\tiny, text=badc, below=0pt of he] {own capability};
  \node[font=\tiny, below=0pt of fo] {own capability};
  \node[c] (pa) at (6.25,1.85) {\cls{Payload}};
  \node[c] (hp) at (6.25,1.0) {\cls{Hazardous-}\\\cls{Payload}};
  \node[new] (ch) at (6.25,0.0) {\cls{Chemical}};
  \draw (pa) -- (hp); \draw (hp) -- (ch);
  \node[font=\tiny, below=0pt of ch] {test only};
\end{tikzpicture}}\\[2pt]
{\small (b) Robot and payload taxonomies}
\end{minipage}
\caption{(a) OntoCBF. (b) Robot classes with per-axis acceleration limits $\bar a$ (m/s$^2$), and payload classes. \cls{LightDrone} (green) is a dynamic subtype of \cls{Drone}. \cls{HeavyDrone} (red) brakes less strongly and \cls{Forklift} has no design ancestor, so both receive their own capability. These classes and \cls{Chemical} appear only in tests.}
\label{fig:overview}
\end{figure}

We let a domain ontology, of the kind already standardized for robotics \citep{ieee1872}, carry the safety specification.
A filter for a heterogeneous team can miss a constraint for a new entity, overestimate the braking capability of a robot, or contain constraints that are each satisfiable and exclude each other.
In our warehouse ontology (Fig.~\ref{fig:overview}), the taxonomy states that a \cls{LightDrone} is a \cls{Drone} and that every \cls{Drone} is \cls{Aerial}.
A safety axiom states how far an \cls{Aerial} agent stays from a \cls{Human}, and a property chain relates each carrier to the people near its payload.
Entailment removes the first failure, because the constraints of unlisted classes, payloads and combinations follow from these statements.
The second failure depends on dynamics, because a \cls{HeavyDrone}, a \cls{Drone} that brakes less strongly, inherits every constraint of \cls{Drone} and needs a braking capability of its own.
A test on the dynamics profiles in the ontology, together with classification, decides when a subclass may reuse the capabilities, and with them the certificate, of its parent.
We prove that the third failure is intrinsic to pairwise barriers.

Our contributions are the following.
\begin{enumerate}\setlength{\itemsep}{1pt}\setlength{\parskip}{0pt}
  \item We formulate safe control of heterogeneous teams whose active constraints are entailed by an ontology and switch online. Our method, OntoCBF, compiles these entailments into a CBF filter with one quadratic program (QP) per agent and trains a shared multi-agent PPO (MAPPO) policy with the filter in the loop.
  \item Theorem~\ref{thm:inherit} shows that a subclass inherits exactly the constraints of its parent under a classification test, and its capabilities under dynamic subtyping, an inclusion of velocity sets that braking barriers of box-bounded double integrators require.
  \item Propositions~\ref{prop:switch}--\ref{prop:three} show that, with capabilities assigned this way, safety is lost only after a conflict, and that some states with positive pairwise barriers lead to a collision under every input, for three robots and, under any fixed capability reduction, for large enough teams.
  \item Experiments in a simulated warehouse with unseen classes and up to 32 robots show that OntoCBF removes the first two failures, that every remaining violation follows a conflict, and that the learned policy lowers the rate of infeasible QPs by \res{infRedMin}--\res{infRedMax}\,\% (Section~\ref{sec:experiments}).
\end{enumerate}

\section{Related Work}\label{sec:related}

\paragraph{Multi-agent safety filters.}
Our barriers and responsibility split follow \citet{wang2016hetero,wang2017barrier}, and \citet{kim2026cahcbf} extend the split to kinematically heterogeneous robots.
\citet{sah2026feasibility} give an exact pointwise test of joint feasibility for heterogeneous teams.
Learned graph CBFs enforce collision avoidance in teams of any size \citep{zhang2025gcbfplus}, are trained jointly with PPO-style policies \citep{zhang2025dgppo}, and now cover heterogeneous teams through type embeddings \citep{yang2026hgcbf} or, for unseen robot models, through features of their reachable velocity sets \citep{kim2026cchgcbf}.
Constrained MARL bounds expected costs \citep{gu2023safemarl}, and shields from temporal-logic specifications act on discrete abstractions \citep{elsayedaly2021shield,xiao2023dynamic}.
Several works check or design the compatibility of multiple CBFs under input limits \citep{tan2022compatibility,breeden2023compositions,isaly2024feasibility,mousavi2025vertices}, and Proposition~\ref{prop:three} shows that pairwise braking barriers are incompatible on parts of their own safe set.

\paragraph{Semantic safety and reuse of certificates.}
Semantic safety filters derive constraints from language or vision-language models \citep{brunke2025semantic,ravichandran2026core}, and logical specifications have been compiled into CBFs as compositions of state constraints \citep{lindemann2019stl,molnar2023composing}.
Ontologies model safety in human--robot collaboration \citep{hrcsafety2026} and have been used to verify robot plans \citep{wu2026virf}, to shape risk fields for motion planning \citep{vanderploeg2023dots} and to mask the actions of a learner \citep{fatima2026triage}.
To our knowledge, OntoCBF is the first method that compiles entailments into control constraints with an invariance guarantee.
Behavioral subtyping requires a subtype to preserve what clients of the supertype rely on \citep{liskov1994}.
A system that admits at least the inputs of another can reuse its controllers \citep{reissig2017frr}, and barriers transfer between systems through reduced-order models \citep{cohen2024rom}, learned inverse dynamics \citep{nadali2024inverse} or a simulation function with a margin \citep{bousias2026transfer}.
Part~2 of Theorem~\ref{thm:inherit} is the zero-margin, class-level form of such a transfer.
Parts~1 and~3 add what an ontology requires, namely a classification test for identical constraints and, for box-bounded double integrators, the necessity of dynamic subtyping for braking barriers (Table~\ref{tab:related}).

\begin{table}[t]
\centering\small
\caption{Related work compared on team certificate, class-specific dynamics, typed relational constraints, formal specification, guarantee across switches, unseen types without redesign, certificate-reuse condition and learning in the loop (\cmark\ addressed, \pmark\ partly, \xmark\ absent).}
\label{tab:related}
\resizebox{\linewidth}{!}{%
\begin{tabular}{l cccccccc}
\toprule
Line of work & Team & Het. & Sem. & Spec. & Switch & New & Reuse & Learn\\
\midrule
Multi-robot CBF filters \citep{wang2017barrier,kim2026cahcbf,sah2026feasibility} & \pmark & \cmark & \xmark & \xmark & \pmark & \xmark & \xmark & \xmark\\
Learned graph CBFs \citep{zhang2025gcbfplus,yang2026hgcbf,kim2026cchgcbf} & \pmark & \cmark & \xmark & \xmark & \pmark & \pmark & \xmark & \cmark\\
Safe MARL and shields \citep{gu2023safemarl,elsayedaly2021shield,chatterji2025smarl} & \pmark & \xmark & \pmark & \pmark & \xmark & \xmark & \xmark & \cmark\\
Logic compiled into CBFs \citep{lindemann2019stl,molnar2023composing} & \pmark & \xmark & \pmark & \cmark & \cmark & \xmark & \xmark & \xmark\\
Constraints from (V)LMs \citep{brunke2025semantic,ravichandran2026core} & \xmark & \xmark & \cmark & \xmark & \pmark & \pmark & \xmark & \pmark\\
Ontologies for robot safety \citep{wu2026virf,vanderploeg2023dots,fatima2026triage} & \xmark & \xmark & \cmark & \cmark & \xmark & \pmark & \xmark & \pmark\\
Certificate transfer \citep{reissig2017frr,bousias2026transfer} & \xmark & \cmark & \xmark & \xmark & \xmark & \pmark & \cmark & \xmark\\
\midrule
\textbf{OntoCBF (ours)} & \pmark & \cmark & \cmark & \cmark & \cmark & \cmark & \cmark & \cmark\\
\bottomrule
\end{tabular}}
\end{table}

\section{Problem Setting}\label{sec:problem}

\paragraph{Entities and dynamics.}
The workspace contains controlled agents $\mathcal{I}=\{1,\dots,N\}$, humans $\mathcal{H}$, restricted zones $\mathcal{Z}$ and payloads $\mathcal{P}$.
Each entity $e$ belongs to a class $\kappa(e)$, a concept name of the ontology, and we call the restriction of $\kappa$ to $\mathcal{I}$ the \emph{team composition}.
An agent of class $C$ has dynamics $\dot x_i=f_C(x_i)+g_C(x_i)u_i$ with $u_i\in U_C$ compact.
In our experiments it is a planar double integrator, $\dot p_i=v_i$, $\dot v_i=u_i$, with $U_C=[-\bar a_C,\bar a_C]^2$.
Humans are double integrators with unknown accelerations bounded by $\|\dot v_h\|\le\bar a_h$, an agent carries at most one payload, which sits at $p_i+o_i$ for a fixed offset $o_i$, and zones are static.

\paragraph{Ontology.}
We use the description logic of OWL~2 \citep{baader2003dl}, in which an ontology consists of a TBox $\T$ of terminological axioms and an ABox $\A$ of facts about individuals.
Our TBox contains the taxonomy of Fig.~\ref{fig:overview}b as concept inclusions $C\sqsubseteq C'$, disjointness axioms such as $\cls{Aerial}\sqcap\cls{Ground}\sqsubseteq\bot$, the symmetric role $\mathrm{near}$, the role $\mathrm{carries}$ from \cls{ControlledAgent} to \cls{Payload}, and the role $\mathrm{nearCarried}$, which the property chain $\mathrm{carries}\circ\mathrm{near}\sqsubseteq\mathrm{nearCarried}$ derives from $\mathrm{carries}(a,p)$ and $\mathrm{near}(p,b)$.
Payloads have their own taxonomy, $\cls{Chemical}\sqsubseteq\cls{HazardousPayload}\sqsubseteq\cls{Payload}$, and the defined class $\cls{HazardCarrier}\equiv\cls{ControlledAgent}\sqcap\exists\mathrm{carries}.\cls{HazardousPayload}$ contains whichever agent currently carries a hazardous payload.
Each robot class is annotated with a dynamics profile, here $\bar a_C$ and a nominal speed $\bar v_C$.
We write $\T\cup\A\models\varphi$ if $\varphi$ holds in every model of $\T\cup\A$, which HermiT decides soundly and completely for OWL~2~DL \citep{glimm2014hermit}.

\paragraph{Safety axioms and active constraints.}
A safety axiom $\sigma=(r_\sigma,C_\sigma,C'_\sigma,D_\sigma)$ requires that the distance between $a$ and $b$ stays at least $D_\sigma$ whenever $r_\sigma(a,b)$, $C_\sigma(a)$ and $C'_\sigma(b)$ are entailed.
For $r_\sigma=\mathrm{nearCarried}$ the distance is measured from the payload of $a$.
The ABox at time $t$,
\begin{equation}\label{eq:abox}
\A_t=\{\kappa(e)(e)\}_{e}\ \cup\ \{\mathrm{carries}(i,p)\}_{i\text{ carries }p}\ \cup\ \{\mathrm{near}(e,e'):\ e\neq e',\ \|p_e-p_{e'}\|<\rho\},
\end{equation}
changes when an entity enters or leaves the sensing radius $\rho$ of another and when an agent picks up or drops a payload.
The constraints active at time $t$ are the entailed instances of the safety axioms,
\begin{equation}\label{eq:active}
\K_t=\big\{(\sigma,a,b):\ \T\cup\A_t\models r_\sigma(a,b)\wedge C_\sigma(a)\wedge C'_\sigma(b)\big\},
\end{equation}
and the safe set is $\mathcal{S}_t=\{x:\ \|q_{\sigma,a}(x)-p_b\|\ge D_\sigma\text{ for all }(\sigma,a,b)\in\K_t\}$, where $q_{\sigma,a}$ is the position of $a$, or of its payload if $r_\sigma=\mathrm{nearCarried}$.
Since a subclass inherits every axiom attached to its ancestors, the safe set depends on the team composition through entailment.

\paragraph{Barrier functions.}
For a constraint $k=(\sigma,a,b)$ let $\Delta p=q_{\sigma,a}-p_b$, $\Delta v=v_a-v_b$, $d=\|\Delta p\|$ and $n=\Delta p/d$.
We use the braking barrier of \citet{wang2016hetero},
\begin{equation}\label{eq:barrier}
h_k=\sqrt{2A_k\,(d-D_\sigma)}+n^\top\Delta v,
\end{equation}
where $A_k>0$ is the braking capability available to the pair.
If $h_k\ge0$, braking at rate $A_k$ along $n$ stops the approach at a distance of at least $D_\sigma$, and $h_k$ is continuously differentiable wherever $d>D_\sigma$.
Since $\dot h_k$ is affine in the accelerations, the condition $\dot h_k\ge-\alpha(h_k)$ with $\alpha(h)=\gamma h^3$ reads
\begin{equation}\label{eq:cbfrow}
n^\top(\dot v_a-\dot v_b)\ \ge\ b_k:=-\gamma h_k^3-\frac{A_k\,n^\top\Delta v}{\sqrt{2A_k(d-D_\sigma)}}-\frac{\|\Delta v\|^2-(n^\top\Delta v)^2}{d}.
\end{equation}

\paragraph{Velocity sets and certificates.}
For a class $C$, $V_C(x_i)=\{f_C(x_i)+g_C(x_i)u:\ u\in U_C\}$ is the set of state derivatives, called velocities as in viability theory, that an agent of class $C$ can produce at $x_i$.
For a team composition $\kappa$ let $V_\kappa(x)=\prod_{i\in\mathcal{I}}V_{\kappa(i)}(x_i)$, and let $\Omega(x)=\prod_{h\in\mathcal{H}}\{(v_h,w):\|w\|\le\bar a_h\}$ be the set of admissible velocities of the humans.

\begin{definition}[Dynamic subtype]\label{def:dyn}
A robot class $C'$ is a \emph{dynamic subtype} of a robot class $C$, written $C'\dyn C$, if both classes have the same state space and $V_C(x_i)\subseteq V_{C'}(x_i)$ for all $x_i$.
\end{definition}

\begin{definition}[Joint validity]\label{def:joint}
A family $\{h_k\}_{k\in\K}$ of functions that are continuously differentiable on an open set containing $\X$ is \emph{jointly valid} for $\kappa$ on $\X$ if for every $x\in\X$ there is $\xi\in V_\kappa(x)$ such that $\nabla_{\mathcal{I}}h_k(x)^\top\xi+\nabla_{\mathcal{H}}h_k(x)^\top\omega\ge-\alpha(h_k(x))$ for all $k\in\K$ and all $\omega\in\Omega(x)$.
\end{definition}

Subsumption $C'\sqsubseteq C$ is entailed by the TBox, and $C'\dyn C$ is a property of the dynamics profiles.
For box-bounded double integrators $C'\dyn C$ holds if and only if $\bar a_{C'}\ge\bar a_C$, so $\cls{HeavyDrone}\sqsubseteq\cls{Drone}$ holds and $\cls{HeavyDrone}\dyn\cls{Drone}$ fails.
A \emph{certificate} for a team is a family that is jointly valid on a set $\X$, together with the capabilities and the gain $\gamma$ used to build it.

\paragraph{Objective.}
Each agent is rewarded for reaching a sequence of goals and acts on a local observation.
We seek decentralized policies and a filter that maximize the expected team return subject to $x(t)\in\mathcal{S}_t$ for all $t$, for any team and any classes that the ontology declares, including those absent at design time and in training.

\section{Compiling the Ontology into a Safety Filter}\label{sec:method}

OntoCBF consists of a reasoner that decides which constraints are active, a CBF filter that enforces them with one QP per agent, and a shared policy trained with MARL that performs the task (Fig.~\ref{fig:overview}a).

\subsection{Compilation and capability assignment}\label{sec:compile}
Offline, HermiT classifies $\T$ once, and OntoCBF stores the ancestors of every class, the safety axioms indexed by role, the defined classes and the dynamics profiles.
The \emph{design classes}, here \cls{Drone}, \cls{Rover} and \cls{Carrier}, are the robot classes of the team for which a certificate is established when the system is designed, and each enters the barriers with the braking capability $\beta_C=\bar a_C$.
Every other robot class $C$ inherits $\beta_C=\beta_{C^+}$ from its nearest design ancestor $C^+$ if $C$ is a dynamic subtype of $C^+$ (Definition~\ref{def:dyn}) and has the same subsumers as $C^+$ in $\Sigma_\T$ (Section~\ref{sec:theory}), a test that classification decides.
By Theorem~\ref{thm:inherit}, $C$ then also inherits the certificate of $C^+$, provided both have the same payload offset.
With several nearest design ancestors, both conditions must hold for each, and the smallest capability is used.
All remaining robot classes, \cls{Forklift} among them, receive $\beta_C=\bar a_C$ and need a certificate of their own.
For box-bounded double integrators the dynamic-subtype test reduces to $\bar a_C\ge\bar a_{C^+}$, so \cls{LightDrone} inherits from \cls{Drone} and \cls{HeavyDrone} receives its own capability.
The rule yields $\beta_C\le\bar a_C$ for every class, which makes every constraint satisfiable on its own wherever its barrier is nonnegative (Lemma~\ref{lem:single}).
Joint validity of all constraints (Definition~\ref{def:joint}) is a separate property, measured in Section~\ref{sec:capability}.

\subsection{Event-triggered reasoning}\label{sec:reasoning}
The reasoner runs at ABox events, which occur when a sensed $\mathrm{near}$ assertion appears or disappears and when the task layer asserts or retracts $\mathrm{carries}$ at a pickup or a drop.
It closes the asserted roles under symmetry and the property chain, realizes the defined classes, so that an agent carrying a hazardous payload becomes a \cls{HazardCarrier}, and instantiates every safety axiom whose role and classes are entailed for a pair.
Each instance becomes a constraint $k$ with keep-out $D_\sigma$, braking capability $A_k$ and a split of \eqref{eq:cbfrow} among its controlled agents.
For two robots, $A_k=\beta_a+\beta_b$, and \eqref{eq:cbfrow} is split into $n^\top u_a\ge w_a b_k$ and $-n^\top u_b\ge w_b b_k$ with $w_a=\beta_a/A_k$ and $w_b=\beta_b/A_k$ \citep{wang2016hetero}.
For a robot and a human, who may accelerate toward the robot, $A_k=\beta_a-\bar a_h$, which is positive for all our classes, and the robot alone enforces $n^\top u_a\ge b_k+\bar a_h$.
For a robot and a zone, $A_k=\beta_a$, and the robot enforces $n^\top u_a\ge b_k$.
We call these linear inequalities the \emph{barrier inequalities} of an agent.
Rule saturation is complete for our TBox, agrees with HermiT on \res{testTrials} random ABoxes and takes \res{tCompiled}\,ms per ABox against \res{tHermit}\,ms.
It also detects inconsistent ABoxes, which would leave $\mathcal{S}_t$ empty, and OntoCBF then keeps the last consistent ABox and brakes (Appendix~\ref{app:dl}).
Reasoning is centralized over the shared ABox, and each QP needs the states and capabilities of the entities within the sensing radius.
\emph{Activation check.} A relation asserted by the task layer, such as $\mathrm{carries}$ at a pickup, can activate a constraint with $h_k<0$, so the reasoner computes the constraints that the assertion would activate and defers the pickup until each has $h_k\ge\res{actEps}$\,m/s.

\subsection{Per-agent filter}\label{sec:filter}
Agent $i$ collects its barrier inequalities as half-planes $g_k^\top u\ge c_k$ and solves
\begin{equation}\label{eq:qp}
u_i=\arg\min_{u\in U_{\kappa(i)}}\ \|u-u_i^{\mathrm{nom}}\|^2\quad\text{s.t.}\quad g_k^\top u\ge c_k\ \ \text{for every barrier inequality $k$ of agent $i$},
\end{equation}
over its true input set $U_{\kappa(i)}$, so the assigned capability enters only through the barriers.
Since $u\in\mathbb{R}^2$, \eqref{eq:qp} is solved exactly over the edges and vertices of the feasible polygon.
If the polygon is empty, the agent minimizes $\|u-u_i^{\mathrm{nom}}\|^2+\mu\sum_k\max(0,c_k-g_k^\top u)^2$ over $U_{\kappa(i)}$ with $\mu=\res{mu}$.
The filter evaluates \eqref{eq:barrier} with $D_\sigma$ enlarged by a margin $\delta=\res{margin}$\,m and requires the relative acceleration $A_k$ along $n$ inside the margin.
We log the single-constraint failures and conflicts that Section~\ref{sec:theory} distinguishes.
A \emph{single-constraint failure} is a barrier inequality with $h_k\ge0$ that every input in $U_{\kappa(i)}$ violates, and a \emph{conflict} is an empty polygon whose inequalities are each satisfiable on their own.

\subsection{Learning with the filter in the loop}\label{sec:learning}
A policy $\pi_\theta$ shared by all agents is trained with MAPPO \citep{yu2022mappo} in the design scenario W0 only, with the filter acting on its outputs.
Agent $i$ observes its own state and goal and, for every entity within $\rho$, the relative state, a class embedding and the keep-out distance that the reasoner currently entails for the pair, pooled by attention.
Class embeddings are tied along the taxonomy, $e_C=\sum_{C'\sqsupseteq C}\delta_{C'}$ with $\delta_{C'}=0$ for classes absent from training, so an unseen class is represented through its known ancestors (Appendix~\ref{app:train}).
The policy outputs a bounded correction $u_i^{\mathrm{nom}}=\mathrm{clip}(u_i^{\mathrm{PD}}+s_i\tanh(a_i))$ to a go-to-goal controller, with $s_i$ capped at the authority seen in training, a cap introduced after an earlier experiment with \cls{LightDrone} (Appendix~\ref{app:train}).
The reward is goal progress plus an arrival bonus minus penalties on the intervention $\|u_i-u_i^{\mathrm{nom}}\|^2$, on infeasible QPs, on violations and on overspeed, with the gradient taken with respect to the pre-filter action $a_i$.
By Proposition~\ref{prop:conflict}, in continuous time the penalty on infeasible QPs targets the only failure mode left under OntoCBF.

\section{Inheritance and Conflicts}\label{sec:theory}

Throughout this section and its proofs in Appendix~\ref{app:proofs}, the agent set $\mathcal{I}$ is fixed, uncontrolled entities keep their classes, and $\T$ is the TBox of Appendix~\ref{app:dl}.

\begin{lemma}[Single-constraint feasibility]\label{lem:single}
Let agent $a$ have the input set $[-\bar a_a,\bar a_a]^2$ and capability $\beta_a$, and consider one of its barrier inequalities from Section~\ref{sec:reasoning}.
\begin{enumerate}
  \item If $\beta_a\le\bar a_a$, then at every state with $d>D_\sigma$ and $h_k\ge0$ some input of $a$ satisfies it.
  \item If $\beta_a>\bar a_a$, there is a state with $h_k=0$ at which every input of $a$ violates the inequality, and for a zone, or a human who accelerates toward the robot, there is a state with $h_k>0$ from which every admissible input of $a$ leads into the keep-out region.
\end{enumerate}
\end{lemma}

Part~1 follows from $b_k\le A_k$ on $\{h_k\ge0\}$, and part~2 from a head-on approach along a coordinate axis.
Let $\Sigma_\T$ be the set of concept names that occur in safety axioms or in TBox axioms other than inclusions between two concept names.

\begin{theorem}[Certificate inheritance]\label{thm:inherit}
Let the team composition $\kappa'$ differ from $\kappa$ on $J\subseteq\mathcal{I}$, and for every $i\in J$ let $\T\models\kappa'(i)\sqsubseteq\kappa(i)$, let $\kappa'(i)$ and $\kappa(i)$ have the same subsumers in $\Sigma_\T$, the same payload offset and the same capability, $\beta_{\kappa'(i)}=\beta_{\kappa(i)}$.
\begin{enumerate}
  \item For all $\mathrm{near}$ and $\mathrm{carries}$ relations, $\kappa$ and $\kappa'$ yield the same constraints, barriers and weights.
  \item If $\kappa'(i)\dyn\kappa(i)$ for all $i\in J$, every family that is jointly valid for $\kappa$ on $\X$ is jointly valid for $\kappa'$ on $\X$, and every QP~\eqref{eq:qp} that is feasible under $\kappa$ is feasible under $\kappa'$.
  \item For double integrators with box input sets and $\beta_{\kappa(i)}=\bar a_{\kappa(i)}$, every barrier inequality of agent $i\in J$ is feasible at all states with $d>D_\sigma$ and $h_k\ge0$ if and only if $\kappa'(i)\dyn\kappa(i)$. If $\kappa'(i)\ndyn\kappa(i)$, the zero-superlevel set of an inherited zone or human barrier contains states from which every admissible input of agent $i$ leads into the keep-out region.
\end{enumerate}
\end{theorem}

\noindent\emph{Proof sketch.}
Entailment is monotone and equal subsumers in $\Sigma_\T$ add no axiom instance (Lemma~\ref{lem:sigma}), so the inequalities are unchanged, and any witness or feasible QP input for $\kappa$ remains one for $\kappa'$ because the velocity sets grow.
Part~3 is Lemma~\ref{lem:single} with capability $\bar a_{\kappa(i)}$ and bound $\bar a_{\kappa'(i)}$.\hfill$\square$

Part~3 shows that, for box-bounded double integrators, the capability of the parent keeps every inherited barrier inequality feasible exactly when the subclass is also a dynamic subtype.
Lemma~\ref{lem:sigma} applies to any entity, so a payload subclass with the same subsumers in $\Sigma_\T$, such as \cls{Chemical}, leaves the active constraints unchanged.
Typing a \cls{Payload} as \cls{HazardousPayload} changes them, since its carrier becomes a \cls{HazardCarrier}.
Appendix~\ref{app:proofs} extends the theorem to other state spaces through a $C^1$ map $\pi$ with $V_C(\pi(x'))\subseteq D\pi(x')V_{C'}(x')$, as for a unicycle, and to inclusions up to a margin.

\begin{assumption}[Activation margin]\label{ass:margin}
Every constraint $k$ entering $\K_t$ at time $t_k$ has $h_k(x(t_k))>0$.
\end{assumption}

As relations switch, Assumption~\ref{ass:margin} holds for sensed relations if closing speeds are bounded by $\bar v$ and $\rho>D_\sigma+\delta+\bar v^2/(2A_k)$, because a constraint activates at distance $\rho$ (Lemma~\ref{lem:radius}).
At nominal speeds our pairs need at most \res{rhoNeed}\,m, for a \cls{\res{rhoNeedCls}} and a human, and we use $\rho=7$\,m.
For relations asserted by the task layer, the activation check enforces it.

\begin{proposition}[Invariance under switching]\label{prop:switch}
Suppose that $\K_t$ changes finitely often on bounded intervals, that Assumption~\ref{ass:margin} holds, and that $h_k(x(0))>0$ for all $k\in\K_0$.
If the inputs applied along an absolutely continuous closed-loop trajectory satisfy every barrier inequality of every agent for almost every $t\in[0,T]$, then $x(t)\in\mathcal{S}_t$ for all $t\in[0,T]$.
\end{proposition}

The proof applies the comparison lemma between switches, which keeps every active $h_k$ positive and every distance above the keep-out, where $h_k$ is differentiable.

\begin{proposition}[Loss of safety starts with a conflict]\label{prop:conflict}
Let the first three hypotheses of Proposition~\ref{prop:switch} hold, let the closed-loop trajectory be absolutely continuous, let every class satisfy $\bar a_h<\beta_C\le\bar a_C$, and let every agent apply the solution of its QP whenever that QP is feasible.
Suppose that the state first leaves the zero-superlevel set of an active barrier at $t^\circ$.
Then some constraint $k$, active on $[t_k,t^\circ]$ with $t_k<t^\circ$, has $h_k(x(t^\circ))=0$, and some agent of $k$ has an infeasible QP on a subset of $[t_k,t^\circ)$ of positive measure, with each of its barrier inequalities feasible on its own at those times.
\end{proposition}

Before $t^\circ$ every active barrier is nonnegative, so Lemma~\ref{lem:single} makes each inequality feasible, and an infeasible QP can only come from inequalities that exclude each other.
Such conflicts are a property of pairwise barriers, independent of the ontology.

\begin{proposition}[Unavoidable conflicts]\label{prop:three}
Consider three double integrators with input sets $[-\bar a,\bar a]^2$, capabilities $\beta=\bar a$ and the robot--robot barriers \eqref{eq:barrier} with $A_k=2\bar a$ and keep-out $D$.
There is a state with all three barriers positive from which every admissible input of the robots brings two of them closer than $D$.
\end{proposition}

Two robots approach a third from opposite sides, and each pairwise barrier assumes that the middle robot devotes its full braking to that pair.
Hence no controller renders the zero-superlevel set of pairwise braking barriers forward invariant, and a certificate for them holds only on a strict subset $\X$ of it, which makes precise the remark of \citet{wang2017barrier} that their QP may become infeasible in crowded scenes.
A chain of $m$ robots extends the construction to capabilities $\beta=\theta\bar a$ whenever $\theta(m-1)>1$, so every fixed capability reduction leaves such conflicts in large enough teams.

\section{Experiments}\label{sec:experiments}

\paragraph{Setup.}
Each $12\times12$\,m warehouse tile (Appendix~\ref{app:exp}) contains two drones that carry parcels across a restricted zone binding only \cls{Ground} agents, four carriers with payloads, two of them hazardous, and two patrolling rovers.
Its four humans ignore the robots, walk at up to 0.8\,m/s and accelerate at up to 0.5\,m/s$^2$.
Besides the robot--robot (0.3\,m), robot--human (0.5\,m), aerial--human (1.0\,m) and ground--zone (1.0\,m) axioms, payloads keep 0.8\,m and hazardous payloads 2.0\,m from people.
Episodes last 40\,s with a zero-order hold of 0.05\,s, whose inter-sample effect we do not bound \citep{breeden2022sampled}, $\rho=7$\,m, and $\gamma=\res{gammaMain}$ is selected on validation episodes.
W0 is the design team (\cls{Drone}, \cls{Rover} and \cls{Carrier} with \cls{Payload} and \cls{HazardousPayload}).
W1 replaces the two drones by a \cls{LightDrone} and a \cls{HeavyDrone} and one carrier by a \cls{Forklift}.
W2 types the hazardous payloads as \cls{Chemical}, W3 lets a \cls{LightDrone} carry a \cls{Chemical} and a \cls{Drone} carry a \cls{HazardousPayload}, two unlisted combinations, and W4 tiles W0 $2\times2$ ($N=32$).

\paragraph{Methods and metrics.}
All filters share the barrier \eqref{eq:barrier}, the QP \eqref{eq:qp} and, unless stated otherwise, the activation check, so they differ only in where their constraints and capabilities come from.
\emph{None} applies $u_i^{\mathrm{nom}}$ directly.
\emph{Flat} replaces the ontology by hand-written tables of keep-outs and capabilities for the design classes and (robot, payload) pairs, with defaults for unlisted classes (mean design capability \res{flatAlpha}\,m/s$^2$, keep-outs \res{flatHumanD}\,m and \res{flatPayD}\,m), and \emph{Flat-cons} uses the most conservative listed values as defaults.
\emph{No-chain} drops property chains, \emph{Naive} inherits capabilities along $\sqsubseteq$ without the dynamic-subtype test, \emph{Untyped} treats every payload as hazardous, and \emph{No-act.} omits the activation check.
Covering W1--W3 requires \res{effOnto} ontology statements in OntoCBF and \res{effFlat} table entries in Flat, and all methods use the same go-to-goal PD controller, except MAPPO, on \res{nSeeds} seeds per scenario.
A violation is a distance below $D_\sigma$ for a constraint entailed under the true classes, so Flat and No-chain are also judged on the constraints they omit.

\begin{table}[t]
\centering\small
\caption{Episodes with a violation (left, 100\,\% for None everywhere) and single-constraint failures summed over episodes (right, zero for all other methods), \res{nSeeds} seeds per cell, all with the PD controller except MAPPO, which runs behind OntoCBF (\res{nRuns} training runs pooled).}
\label{tab:safety}
\resizebox{0.97\linewidth}{!}{\begin{tabular}{l rrrrrrrr rrr}
\toprule
 & \multicolumn{8}{c}{Episodes with a violation (\%)} & \multicolumn{3}{c}{Single-constraint failures} \\
\cmidrule(lr){2-9}\cmidrule(lr){10-12}
Scenario & Flat & Flat-cons & No-chain & Naive & Untyped & No-act. & \textbf{OntoCBF} & MAPPO & Flat & Naive & \textbf{OntoCBF} \\
\midrule
W0 (design) & 20 & 20 & 100 & 22 & 16 & 74 & 22 & 12 & 0 & 0 & 0 \\
W1 (unseen robots) & 89 & 17 & 100 & 23 & 17 & 72 & 20 & 11 & 182 & 266 & 0 \\
W2 (unseen payload) & 99 & 20 & 100 & 22 & 16 & 74 & 22 & 12 & 0 & 0 & 0 \\
W3 (unseen combination) & 100 & 28 & 100 & 40 & 37 & 96 & 40 & 26 & 0 & 0 & 0 \\
W4 ($N{=}32$) & 67 & 67 & 100 & 70 & 69 & 98 & 70 & 65 & 0 & 0 & 0 \\
\bottomrule
\end{tabular}
}
\end{table}

\subsection{Missing constraints}\label{sec:zeroshot}
OntoCBF runs unchanged in all scenarios with the same safety axioms, and W2 reproduces W0 seed by seed, as Lemma~\ref{lem:sigma} predicts.
In W1 the \cls{HeavyDrone} and the \cls{Forklift} run with their own capabilities, for which a certificate is still to be established.
Under OntoCBF, \res{vpOntoMin}--\res{vpOntoMax}\,\% of the episodes contain a violation, and an episode has \res{vsOntoMin}--\res{vsOntoMax} violating steps out of 800 on average, against 100\,\% and \res{vsNoneMin}--\res{vsNoneMax} steps without a filter (Tables~\ref{tab:safety} and~\ref{tab:learned}, intervals in Table~\ref{tab:ci}).
Flat performs like OntoCBF on the design team (\res{vp/W0/Flat}\,\% against \res{vp/W0/Onto}\,\%, where OntoCBF keeps one inequality per entailed axiom and Flat one per table entry) and violates safety in at least \res{vpFlatUnseenMin}\,\% of the episodes of W1--W3, mostly through the omitted aerial--human axiom in W1 (\res{ax/W1/Flat/air} violating constraint-steps per episode against \res{ax/W1/Onto/air}) and the omitted hazard keep-out in W2 and W3.
Flat-cons, which treats every unlisted robot or pair like the weakest and most restricted listed one, violates safety less often than OntoCBF in every scenario, most of all in W3 (\res{vp/W3/FlatCons} against \res{vp/W3/Onto}\,\%).
Its zone default stops the unseen drones, so it achieves \res{fcDelW1} and \res{fcDelW3}\,\% of the deliveries of OntoCBF in W1 and W3.
Its capability default is valid only because every unseen class brakes at least as strongly as the weakest design class (Lemma~\ref{lem:single}).
Untyped has fewer episodes with a violation in every scenario and achieves \res{untypedOrdPct}\,\% of the ordinary deliveries of OntoCBF in W0.
Removing the property chain, the robot or payload taxonomy, or the defined class produces violations in at least \res{singleMinVp}\,\% of the episodes where the element is needed and reproduces OntoCBF seed by seed elsewhere (Tables~\ref{tab:ingredients} and~\ref{tab:single}).
Without the activation check, \res{act/W3/NoAct} pickups in W3 begin inside a keep-out region (Table~\ref{tab:ingredients}).

\subsection{Overestimated capabilities}\label{sec:capability}
A \cls{HeavyDrone} heads for a goal through a human who walks straight at it, from \res{stN} initial states inside the safe set of its own capability (Fig.~\ref{fig:stress}).
Under Naive, which gives it the capability of \cls{Drone}, single-constraint failures occur in \res{stNaiveCfMain}\,\% of the encounters at $\gamma=\res{gammaMain}$, \res{stCfFirstPct}\,\% at $\gamma=\res{stGcfFirst}$ and \res{stNaiveCf}\,\% at $\gamma=\res{stGmax}$.
From $\gamma=\res{stGfirst}$ on they lead to violations, in \res{stNaiveViol}\,\% of the encounters at $\gamma=\res{stGmax}$ with distances down to \res{stNaiveDmin}\,m against a keep-out of 1.0\,m, as part~3 of Theorem~\ref{thm:inherit} predicts.
OntoCBF stays free of failures and violations at every gain.
Small gains brake early, so an overestimated capability appears as single-constraint failures before it appears as violations.
In W1, Naive causes \res{cf/W1/Naive} single-constraint failures at $\gamma=\res{gammaMain}$ and, at $\gamma=10$, violates safety in \res{gn/10/vpN}\,\% of the episodes against \res{gn/10/vpO}\,\% for OntoCBF (Table~\ref{tab:gamma}).
Joint validity, decided by a linear program at each of the \res{jvStates} states visited in W0 (Table~\ref{tab:joint}), holds for the design family at \res{jvDesignPct}\,\% of them.
\cls{HeavyDrone}s under the capability of \cls{Drone} break it at \res{jvHeavyFail} of these states, and with their own capability they form a new family, valid at \res{jvOwnPct}\,\% of the states where the design family is.

\subsection{Conflicts between constraints}\label{sec:cost}
Each of the \res{vsrcOnsetsDistinct} violation onsets in W0, W1, W3 and W4 follows, within 1\,s, a conflict of an agent of the violated constraint, consistent with Proposition~\ref{prop:conflict} in continuous time, so every episode with a violation also has a conflict.
Conflicts precede only \res{vcBaseLo}--\res{vcBaseHi}\,\% of all agent-steps within the same window (Table~\ref{tab:cause}).
Of all infeasible QPs, \res{vcJointShare}\,\% are conflicts, and the rest contain an inequality whose barrier is already negative.
Conflicts arise mostly when humans close in on a drone or a hazardous carrier, and the median deepest intrusion is \res{depMedLo}--\res{depMedHi}\,\% of $D_\sigma$.
At \res{notCloseLo}--\res{notCloseHi}\,\% of the onsets the robot was not moving toward the other entity.
The filter takes at most \res{tfMax}\,ms per step and the full reasoning step \res{trMax}\,ms per event on one CPU core (Table~\ref{tab:cost}).

\subsection{Learning in the loop}\label{sec:learned}
Behind the MAPPO policy of Section~\ref{sec:learning}, trained in W0 only, episodes with a violation fall from \res{vpOntoMin}--\res{vpOntoMax}\,\% to \res{vpRLMin}--\res{vpRLMax}\,\% (Table~\ref{tab:safety}).
The rate of infeasible QPs per agent-step with active constraints, almost all of them conflicts, drops by \res{infRedMin}--\res{infRedMax}\,\% in every scenario, with paired 95\,\% bootstrap intervals below zero over runs and seeds and in each of the \res{nRuns} runs (Table~\ref{tab:runs}).
Mean violating steps per episode change by \res{dvsMin} to \res{dvsMax}, significantly only in W3 (\res{dvs/W3}), deliveries per courier by \res{ddelMin} to \res{ddelMax}, and without the filter the policy violates safety in every episode.

\section{Conclusion}\label{sec:conclusion}

OntoCBF makes a domain ontology the specification of a multi-robot safety filter, in which entailment removes missing constraints and the dynamic-subtype test, exact for box-bounded double integrators, removes overestimated capabilities.
Conflicts between constraints remain possible with pairwise barriers (Proposition~\ref{prop:three}), and a policy trained in W0 alone makes them less frequent.
The guarantees hold in continuous time for double integrators and humans with acceleration bounded by $\bar a_h$, and they rely on correct class labels and measured profiles, since each entity is constrained only by the axioms of the classes entailed for it (Appendix~\ref{app:dl}).
Verified certificates, fail-safe defaults for unknown labels, axiom priorities that favor people, and nonlinear classes are the next steps.

\subsection*{Generative AI Use}
Generative AI tools were used for coding and experimental implementation, literature search, and writing and editing. The authors checked all AI-assisted content and take full responsibility for the paper.

\bibliography{refs}

\appendix

\section{Proofs}\label{app:proofs}

The constructions in part~2 of Lemma~\ref{lem:single}, part~3 of Theorem~\ref{thm:inherit} and Proposition~\ref{prop:three} show entry into the keep-out used in the barrier, which for the filter is $D_\sigma+\delta$ (Section~\ref{sec:filter}).

\subsection{Barrier derivative and barrier inequalities}\label{sec:hdot}
Let $k=(\sigma,a,b)$ be a constraint with $d=\|\Delta p\|>D_\sigma$, $n=\Delta p/d$ and $\ell=\sqrt{2A_k(d-D_\sigma)}>0$.
The payload offset is constant, so $\tfrac{d}{dt}\Delta p=\Delta v$ for every type of constraint, $\dot d=n^\top\Delta v$, and $\dot n=(I-nn^\top)\Delta v/d$.
Differentiating \eqref{eq:barrier} gives
\begin{equation}\label{eq:hdot}
\dot h_k=\frac{A_k\,n^\top\Delta v}{\ell}+\frac{\|\Delta v\|^2-(n^\top\Delta v)^2}{d}+n^\top(\dot v_a-\dot v_b),
\end{equation}
so $\dot h_k\ge-\gamma h_k^3$ is equivalent to \eqref{eq:cbfrow}.
For two robots, adding $n^\top u_a\ge w_ab_k$ and $-n^\top u_b\ge w_bb_k$ with $w_a+w_b=1$ yields $n^\top(u_a-u_b)\ge b_k$.
For a robot and a human with $\|\dot v_h\|\le\bar a_h$, the inequality $n^\top u_a\ge b_k+\bar a_h$ yields $n^\top(u_a-\dot v_h)\ge n^\top u_a-\bar a_h\ge b_k$ for every admissible acceleration of the human.
For a zone, $\dot v_b=0$.
Hence, whenever the agents of $k$ satisfy their barrier inequalities, $\dot h_k\ge-\gamma h_k^3$ holds for all admissible human accelerations.
The curvature term in \eqref{eq:hdot} is nonnegative, so $h_k\ge0$ is sufficient for braking at rate $A_k$ along $n$ to stop the approach at a distance of at least $D_\sigma$, and tangential motion can only increase that distance.

\subsection{Single-constraint feasibility}

\begin{proof}[of Lemma~\ref{lem:single}]
\emph{Part 1.}
Let $d>D_\sigma$ and $h_k\ge0$.
In $b_k$, the first term $-\gamma h_k^3$ is nonpositive and the third term is nonpositive by the Cauchy--Schwarz inequality.
If $n^\top\Delta v\ge0$, the second term is nonpositive as well.
Otherwise $h_k\ge0$ gives $\ell\ge-n^\top\Delta v>0$, so the second term is at most $A_k$.
Hence $b_k\le A_k$.
The right-hand side $r$ of the barrier inequality of agent $a$ therefore satisfies $r=w_ab_k\le w_aA_k=\beta_a$ for a robot pair, $r=b_k+\bar a_h\le A_k+\bar a_h=\beta_a$ for a human, and $r=b_k\le A_k=\beta_a$ for a zone.
The input $u^\star=\bar a_a(\operatorname{sign}n_1,\operatorname{sign}n_2)$ lies in the box and gives $n^\top u^\star=\bar a_a(|n_1|+|n_2|)\ge\bar a_a\|n\|=\bar a_a\ge\beta_a\ge r$.
Inside the filter margin the implementation replaces $b_k$ by $A_k$, which keeps $r\le\beta_a$.

\emph{Part 2, a state with $h_k=0$.}
Let $\beta_a>\bar a_a$, pick $d>D_\sigma$, and set $n=(1,0)$ and $\Delta v=(-\ell,0)$.
Then $h_k=0$, the curvature term vanishes and $b_k=A_k$, so the right-hand side equals $\beta_a$ for each type of constraint.
Every input in the box violates the inequality, because $\max_{u\in[-\bar a_a,\bar a_a]^2}(1,0)\,u=\bar a_a<\beta_a$.

\emph{Part 2, an unavoidable violation at a zone.}
Write $D=D_\sigma$ and $\bar a=\bar a_a$, place the zone at the origin, and fix $\beta'\in(\bar a,\beta_a)$ and $s>0$.
The agent starts at $p(0)=(D+\varepsilon,0)$ with $v(0)=(-s,0)$ and $\varepsilon=s^2/(2\beta')$, so that $h_k(x(0))=s\big(\sqrt{\beta_a/\beta'}-1\big)>0$.
Let $u(\cdot)$ be any measurable input with values in $[-\bar a,\bar a]^2$, and let $t^\star=s/\bar a$.
Integrating twice,
\begin{equation*}
D+\varepsilon-st-\tfrac12\bar at^2\ \le\ p_1(t)\ \le\ D+\varepsilon-st+\tfrac12\bar at^2,\qquad |p_2(t)|\le\tfrac12\bar at^2 .
\end{equation*}
At $t^\star$ these bounds give $D-\tfrac{3s^2}{2\bar a}\le p_1(t^\star)\le D-cs^2$ with $c=\tfrac12(1/\bar a-1/\beta')>0$, and $|p_2(t^\star)|\le s^2/(2\bar a)$.
If $s^2<\min\{2\bar aD/3,\ D/c,\ 2cD/(c^2+1/(4\bar a^2))\}$, then $0<p_1(t^\star)\le D-cs^2$ and
\begin{equation*}
\|p(t^\star)\|^2\le(D-cs^2)^2+\frac{s^4}{4\bar a^2}=D^2-2cDs^2+\Big(c^2+\frac{1}{4\bar a^2}\Big)s^4<D^2 ,
\end{equation*}
so the agent is inside the keep-out region at $t^\star$ for every admissible input.

\emph{Part 2, an unavoidable violation with a human.}
Let the human start at the origin at rest and accelerate toward the robot with $\dot v_h=(\bar a_h,0)$.
In relative coordinates $\Delta p=p_a-p_h$, the first component of $\Delta\ddot p$ lies in $[-\bar a_a-\bar a_h,\ \hat a]$ with $\hat a=\bar a_a-\bar a_h$, and the second lies in $[-\bar a_a,\bar a_a]$.
The barrier uses $A_k=\beta_a-\bar a_h>\hat a$.
The zone argument with $\beta'\in(\max\{\hat a,0\},A_k)$, $\varepsilon=s^2/(2\beta')$, $t^\star=s/\beta'$, upper acceleration bound $\hat a$, lower bound $-\bar a_a-\bar a_h$ and lateral bound $\bar a_a$ gives $h_k(x(0))>0$ and $\Delta p_1(t^\star)\le D-s^2(1-\hat a/\beta')/(2\beta')$, so $\|\Delta p(t^\star)\|<D$ for all sufficiently small $s>0$, whatever the sign of $\hat a$.
\end{proof}

For the zone, the conditions $s^2\le 2\bar aD\min\{1/3,\ 1-\bar a/\beta'\}$ suffice, and with $\bar a=1$, $\beta_a=2$, $\beta'=1.5$ and $D=1$\,m they allow $s\le0.81$\,m/s.
At $s=0.8$\,m/s the barrier starts at $0.12$\,m/s and the largest reachable distance at $t^\star$ is at most $0.95$\,m.
Smallness of $s$ is needed in this argument, since at large speeds a robot may swerve around the keep-out region.
For a general compact convex input set $U$ containing the origin, the argument of part~1 shows that every barrier inequality is feasible on $\{h_k\ge0\}$ if and only if $\beta_a\le\min_{\|n\|=1}\max_{u\in U}n^\top u$, the inradius of $U$ about the origin.
Inclusion $U_C\subseteq U_{C'}$ implies this inequality for an inherited capability, and joint validity in Theorem~\ref{thm:inherit} needs the inclusion itself, because a joint witness may use a corner of $U_C$.

\subsection{Axiom equivalence and certificate inheritance}

Let $\Sigma_\T$ be the set of concept names that occur in safety axioms or in TBox axioms other than inclusions between two concept names (definitions, disjointness axioms, domain and range restrictions).
For the TBox of Section~\ref{app:dl}, $\Sigma_\T$ consists of \cls{Entity}, \cls{Agent}, \cls{Object}, \cls{ControlledAgent}, \cls{UncontrolledAgent}, \cls{Aerial}, \cls{Ground}, \cls{Human}, \cls{Payload}, \cls{HazardousPayload}, \cls{RestrictedZone} and \cls{HazardCarrier}.
Axioms that tie a class to a role, such as $C\sqsubseteq\exists\mathrm{near}.\{h\}$, put $C$ in $\Sigma_\T$, so a class that adds constraints for other individuals in this way has subsumers in $\Sigma_\T$ that differ from those of its parent.

\begin{lemma}[Axiom equivalence]\label{lem:sigma}
Let $\T$ be the TBox of Section~\ref{app:dl}, and let $C'\sqsubseteq C$ be classes with the same subsumers in $\Sigma_\T$.
Let $\A$ be an ABox of concept-name and role assertions that contains $C(i)$, and let $\A'$ be obtained from $\A$ by replacing $C(i)$ with $C'(i)$.
Then $\T\cup\A'$ is consistent if and only if $\T\cup\A$ is consistent, and both entail the same instances of every safety axiom.
\end{lemma}

\begin{proof}
Since $\T\models C'\sqsubseteq C$, every model of $\T\cup\A'$ is a model of $\T\cup\A$, so $\T\cup\A'$ entails everything that $\T\cup\A$ entails, and consistency of $\T\cup\A'$ implies consistency of $\T\cup\A$.
For the converse, let $\mathcal{U}$ be a model of $\T\cup\A$, and let $\mathcal{U}'$ be obtained from $\mathcal{U}$ by adding $i^{\mathcal{U}}$ to the extension of every concept name $B\notin\Sigma_\T$ with $\T\models C'\sqsubseteq B$.
The ABox $\A$ consists of assertions of concept names and roles, as in \eqref{eq:abox}, and $\mathcal{U}'$ only enlarges extensions, so $\mathcal{U}'$ satisfies $\A$.
It also satisfies $C'(i)$, because $C'$ either received $i^{\mathcal{U}}$ or lies in $\Sigma_\T$, where it is one of its own subsumers and hence a subsumer of $C$.
Every safety axiom and every TBox axiom other than an inclusion between two concept names mentions only individuals, roles and concept names in $\Sigma_\T$, which keep their interpretation in $\mathcal{U}'$.
An inclusion $B\sqsubseteq B'$ between concept names can fail in $\mathcal{U}'$ only if $i^{\mathcal{U}}$ was added to $B$, and then $\T\models C'\sqsubseteq B'$, so $B'$ received $i^{\mathcal{U}}$ as well or $B'\in\Sigma_\T$, in which case $\T\models C\sqsubseteq B'$ and $i^{\mathcal{U}}\in C^{\mathcal{U}}\subseteq {B'}^{\mathcal{U}}$.
Hence $\mathcal{U}'$ is a model of $\T\cup\A'$ with the same safety-axiom instances as $\mathcal{U}$, so $\T\cup\A'$ is consistent whenever $\T\cup\A$ is, and every instance that $\T\cup\A'$ entails is entailed by $\T\cup\A$.
\end{proof}

The proof never uses $C'\sqsubseteq C$ for the converse direction, so any two classes with the same subsumers in $\Sigma_\T$ yield the same constraints.
For example, \cls{Forklift} and \cls{Carrier} are siblings with the same subsumers in $\Sigma_\T$, and \cls{Forklift} has the larger acceleration limit, so by part~2 of Theorem~\ref{thm:inherit}, applied with this weaker hypothesis, the same payload offset and the capability of \cls{Carrier}, it could reuse the certificate of \cls{Carrier}.
OntoCBF searches for a certificate only along the taxonomy, which is conservative.

\begin{proof}[of Theorem~\ref{thm:inherit}]
\emph{Part 1.}
Applying Lemma~\ref{lem:sigma} to the agents of $J$ one at a time shows that, for all $\mathrm{near}$ and $\mathrm{carries}$ relations, the ABoxes under $\kappa$ and under $\kappa'$ entail the same instances of every safety axiom, so $\K_t$ is the same.
A barrier \eqref{eq:barrier} depends on the classes of its agents only through $A_k$, and $A_k$ and the weights depend on the class only through $\beta$, which is the same under $\kappa$ and $\kappa'$.

\emph{Part 2.}
Fix $x\in\X$ and let $\xi\in V_\kappa(x)$ satisfy Definition~\ref{def:joint} for $\kappa$.
For $i\in J$, $\xi_i\in V_{\kappa(i)}(x_i)\subseteq V_{\kappa'(i)}(x_i)$, and for $i\notin J$ the velocity sets coincide, so $\xi\in V_{\kappa'}(x)$.
The functions $h_k$, the index set $\K$, the set $\Omega(x)$ and the value $\alpha(h_k(x))$ do not depend on the composition by part~1, so $\xi$ satisfies Definition~\ref{def:joint} for $\kappa'$.
For the QPs, write each barrier inequality of agent $i$ as $\nabla_{x_i}h_k(x)^\top\big(f_{\kappa(i)}(x_i)+g_{\kappa(i)}(x_i)u\big)\ge s_{ki}(x)$, which reads $n^\top u\ge c_k$ for double integrators.
By part~1, the share $s_{ki}$ of agent $i$ is the same under $\kappa$ and $\kappa'$.
If $u$ is feasible for agent $i\in J$ under $\kappa$, its velocity $f_{\kappa(i)}(x_i)+g_{\kappa(i)}(x_i)u$ lies in $V_{\kappa'(i)}(x_i)$, so some $u'\in U_{\kappa'(i)}$ produces the same velocity and satisfies the same inequalities.
The payload offset enters the barriers of $\mathrm{nearCarried}$ through $q_{\sigma,a}$, which is why the theorem assumes it unchanged.

\emph{Part 3.}
Under $\kappa'$, agent $i$ has capability $\beta=\bar a_{\kappa(i)}$ and input bound $\bar a_{\kappa'(i)}$.
By Lemma~\ref{lem:single}, all its barrier inequalities are feasible at every state with $d>D_\sigma$ and $h_k\ge0$ if $\bar a_{\kappa(i)}\le\bar a_{\kappa'(i)}$, which is the condition $\kappa'(i)\dyn\kappa(i)$, and otherwise each type of inequality is infeasible at some state with $h_k=0$.
The last claim of part~3 is part~2 of Lemma~\ref{lem:single}.
\end{proof}

\paragraph{Feasibility of every QP versus joint validity.}
If every agent's QP is feasible at $x$, the applied inputs satisfy Definition~\ref{def:joint} at $x$, because the two shares of a robot pair sum to the barrier condition and each human inequality contains the worst human acceleration.
The converse fails for two drones with $\beta=\bar a=2$ and $\gamma=1$ on the first axis, drone $b$ at $0$ with velocity $(1,0)$ and drone $a$ at $0.8$ at rest, and a human at $1.914$ at rest.
The robot pair requires $u_{a,1}\ge0.5$ and $-u_{b,1}\ge0.5$, and the aerial--human constraint of $a$ requires $-u_{a,1}\ge0.3$, so the QP of $a$ is infeasible, and each of its inequalities is feasible on its own.
The inputs $u_a=(-0.35,0)$ and $u_b=(-1.4,0)$ satisfy the unsplit pair condition $u_{a,1}-u_{b,1}\ge1$ and both human conditions, so Definition~\ref{def:joint} holds.
On the states visited in W0 this gap separates the \res{jvDesignPct}\,\% of states with joint validity from the \res{jvSplitPct}\,\% at which every QP is feasible (Table~\ref{tab:joint}).

\subsection{Extensions of Theorem~\ref{thm:inherit}}

\paragraph{Classes with different state spaces.}
Let $\pi_i:\mathbb{X}_{\kappa'(i)}\to\mathbb{X}_{\kappa(i)}$, $i\in J$, be continuously differentiable maps between state spaces with
\begin{equation}\label{eq:embed}
V_{\kappa(i)}(\pi_i(x'_i))\subseteq D\pi_i(x'_i)\,V_{\kappa'(i)}(x'_i)\qquad\text{for all }x'_i\in\X'_i,
\end{equation}
and let $\Pi$ apply $\pi_i$ to the states of the agents in $J$ and the identity to all other states.

\begin{proposition}[Inheritance through embeddings]\label{prop:embed}
Let the hypotheses of Theorem~\ref{thm:inherit} hold with $\kappa'(i)\dyn\kappa(i)$ replaced by \eqref{eq:embed}.
If $\{h_k\}_{k\in\K}$ is jointly valid for $\kappa$ on $\X$, then $\{h_k\circ\Pi\}_{k\in\K}$ is jointly valid for $\kappa'$ on $\{x':\ \Pi(x')\in\X,\ x'_i\in\X'_i\text{ for all }i\in J\}$.
\end{proposition}

\begin{proof}
Let $x'$ be in that set and $x=\Pi(x')$, and take a witness $\xi$ for $\kappa$ at $x$.
For $i\in J$ choose $\xi'_i\in V_{\kappa'(i)}(x'_i)$ with $D\pi_i(x'_i)\xi'_i=\xi_i$, which exists by \eqref{eq:embed}, and set $\xi'_i=\xi_i$ for $i\notin J$.
By the chain rule, $\nabla(h_k\circ\Pi)(x')^\top\xi'=\nabla h_k(x)^\top D\Pi(x')\xi'=\nabla h_k(x)^\top\xi$, the terms of the uncontrolled entities are unchanged, and $(h_k\circ\Pi)(x')=h_k(x)$, so $\xi'$ is a witness for $\kappa'$ at $x'$.
\end{proof}

Condition \eqref{eq:embed} is the infinitesimal form of the consistent abstractions of \citet{pappas2000hierarchically}, and it underlies the transfer of CBFs through reduced-order models \citep{cohen2024rom}.
The barrier $h_k\circ\Pi$ constrains the embedded point, so the keep-out of the physical robot must be stated for that point or enlarged accordingly.
The set $\X'_i$ on which \eqref{eq:embed} holds must also be kept invariant, which belongs to the certificate of the design team.

\paragraph{A unicycle inheriting from a double integrator.}
A dynamic unicycle has state $x'=(p,\theta,s,\omega)$ and dynamics $\dot p=s\,e(\theta)$, $\dot\theta=\omega$, $\dot s=a$, $\dot\omega=\eta$, with $|a|\le a_{\max}$, $|\eta|\le\eta_{\max}$, $e(\theta)=(\cos\theta,\sin\theta)$ and $e^\perp(\theta)=(-\sin\theta,\cos\theta)$.
Its look-ahead point $q=p+\ell e(\theta)$ satisfies $\dot q=s\,e(\theta)+\ell\omega\,e^\perp(\theta)$ and
\begin{equation*}
\ddot q=(a-\ell\omega^2)\,e(\theta)+(s\omega+\ell\eta)\,e^\perp(\theta).
\end{equation*}
Let $\pi(x')=(q,\dot q)$.
The position components of $V_{\mathrm{DI}}(\pi(x'))$ and $D\pi(x')V_{\mathrm{uni}}(x')$ coincide, and the inclusion \eqref{eq:embed} of the acceleration components reads
$[-\bar a,\bar a]^2\subseteq R(\theta)\big([-a_{\max}-\ell\omega^2,\,a_{\max}-\ell\omega^2]\times[s\omega-\ell\eta_{\max},\,s\omega+\ell\eta_{\max}]\big)$, where $R(\theta)$ is the rotation by $\theta$.
The squares $R(-\theta)[-\bar a,\bar a]^2$ cover the disk of radius $\sqrt2\,\bar a$ as $\theta$ varies, so the inclusion holds for all $\theta$ on the envelope $|s|\le s_{\max}$, $|\omega|\le\omega_{\max}$ if
\begin{equation*}
a_{\max}-\ell\omega_{\max}^2\ge\sqrt2\,\bar a\qquad\text{and}\qquad \ell\eta_{\max}-s_{\max}\omega_{\max}\ge\sqrt2\,\bar a .
\end{equation*}
These conditions are also necessary for the inclusion at every heading, and on a set of headings $\Theta$ the factor $\sqrt2$ can be replaced by $\max_{\theta\in\Theta}(|\cos\theta|+|\sin\theta|)$.
The barrier then constrains $q$, so the keep-out of the robot itself grows by $\ell$, and the envelope must be kept invariant, for instance through a speed bound $\|\dot q\|\le\bar v$, which implies $|s|\le\bar v$ and $|\omega|\le\bar v/\ell$.
The conditions are demanding, since with $\bar a=1$\,m/s$^2$, $\ell=0.1$\,m, $s_{\max}=1$\,m/s and $\omega_{\max}=2$\,rad/s they need $\eta_{\max}\ge20+10\sqrt2\approx34.1$\,rad/s$^2$.

\paragraph{Approximate inclusion.}
\begin{corollary}\label{cor:robust}
Let the hypotheses of Theorem~\ref{thm:inherit} hold with $\kappa'(i)\dyn\kappa(i)$ replaced by $V_{\kappa(i)}(x_i)\subseteq V_{\kappa'(i)}(x_i)\oplus\mathbb{B}_{\delta_i}$ for $i\in J$.
Suppose that the witnesses for $\kappa$ satisfy Definition~\ref{def:joint} with margin $m>0$ (right-hand side $-\alpha(h_k(x))+m$), and that $|\nabla_{x_i}h_k(x)^\top e|\le L_k\|e\|$ for every difference $e$ between elements of $V_{\kappa(i)}(x_i)$ and $V_{\kappa'(i)}(x_i)$.
If $L_k\sum_{i\in J}\delta_i\le m$ for all $k$, then $\{h_k\}_{k\in\K}$ is jointly valid for $\kappa'$.
\end{corollary}

\begin{proof}
For $i\in J$ pick $\xi'_i\in V_{\kappa'(i)}(x_i)$ with $\|\xi'_i-\xi_i\|\le\delta_i$, and let $\xi'_i=\xi_i$ otherwise.
For every $k$ and $\omega\in\Omega(x)$ the left-hand side of Definition~\ref{def:joint} decreases by at most $L_k\sum_{i\in J}\delta_i\le m$, so it remains at least $-\alpha(h_k(x))$.
\end{proof}

For the barrier \eqref{eq:barrier} and classes with a common drift, the differences only affect accelerations, which enter through $\pm n$, so $L_k=1$.

\subsection{Activation margin from the sensing radius}

\begin{lemma}\label{lem:radius}
Let a constraint $k$ be activated at time $t_k$ by a sensed relation, so that the distance $d$ of $k$ equals $\rho$ at $t_k$.
For $\mathrm{nearCarried}$, the relation $\mathrm{near}$ is sensed at the payload, from which $d$ is measured.
If the closing speed satisfies $-n^\top\Delta v(t_k)\le\bar v$ and $\rho>D_\sigma+\delta+\bar v^2/(2A_k)$, then $h_k(x(t_k))\ge\sqrt{2A_k(\rho-D_\sigma-\delta)}-\bar v>0$.
\end{lemma}

The proof is immediate from \eqref{eq:barrier} with keep-out $D_\sigma+\delta$.
If an implementation senses the relation at the position of the carrier, $\rho$ must exceed the bound by the payload offset $\|o_a\|$.
For the \cls{HeavyDrone} and a human, $\bar a_{\cls{HeavyDrone}}=1.0$, $\bar a_h=0.5$, $D_\sigma=1.0$\,m, $\delta=\res{margin}$\,m and $\bar v=1.5+0.8$\,m/s give $\rho>\res{rhoNeed}$\,m, and at $\rho=7$\,m the barrier is at least $0.13$\,m/s at activation.
A \cls{Drone} or \cls{LightDrone} carrying a hazardous payload needs $3.82$\,m, and a \cls{Carrier} with a hazardous payload needs $4.61$\,m, or $5.31$\,m if the relation is sensed at the carrier.
These values assume nominal speeds, and a \cls{HeavyDrone} at $2.0$\,m/s would need more than $7$\,m, as would a \cls{HeavyDrone} carrying a hazardous payload, a combination outside all our scenarios.
Sampling at $0.05$\,s moves the activation inward by at most about $0.12$\,m, which the slack absorbs.
If $\mathrm{near}$ is retracted only when the distance exceeds $\rho+\eta$ for some $\eta>0$ and relative speeds are bounded by $v_{\max}$, two activations of the same pair are at least $2\eta/v_{\max}$ apart.
Together with a positive dwell time of task events, this guarantees finitely many switches on bounded intervals.

\subsection{Invariance under switching}

\begin{proof}[of Proposition~\ref{prop:switch}]
Let $0=t_0<t_1<\dots<t_m\le T$ be the times in $[0,T]$ at which $\K_t$ changes, and set $t_{m+1}=T$.
For a constraint $k$ write $D_k=D_\sigma+\delta$ for the keep-out used by the filter and $d_k$ for its distance.
We show by induction over $j$ that every $k$ active on $[t_j,t_{j+1})$ satisfies $h_k(x(t))\ge\eta_k(t)>0$ and $d_k(t)>D_k$ for $t\in(t_j,t_{j+1}]$, where $\eta_k$ solves $\dot\eta=-\gamma\eta^3$ with $\eta_k(t_j)=h_k(x(t_j))$.
At $t_j$, a constraint that was active before satisfies $h_k(x(t_j))>0$ by the induction hypothesis and continuity, a newly active one by Assumption~\ref{ass:margin}, and a constraint active at $t_0=0$ by hypothesis.
Fix such a $k$, and let $\tau$ be the supremum of the times $t\in[t_j,t_{j+1}]$ such that $h_k>0$ and $d_k>D_k$ on $(t_j,t]$.
On $(t_j,\tau)$ the function $h_k$ is continuously differentiable along the trajectory, $t\mapsto h_k(x(t))$ is absolutely continuous, and the barrier inequalities of the agents of $k$ hold for almost every $t$, so $\tfrac{d}{dt}h_k(x(t))\ge-\gamma h_k(x(t))^3$ almost everywhere by Section~\ref{sec:hdot}.
The comparison lemma gives $h_k(x(t))\ge\eta_k(t)$ on $[t_j,\tau)$, hence $h_k(x(\tau))\ge\eta_k(\tau)>0$ by continuity.
If $d_k(\tau)=D_k$, then $\dot d_k(\tau)=n^\top\Delta v(\tau)=h_k(x(\tau))>0$, which is impossible because $d_k>D_k$ on $(t_j,\tau)$.
So $d_k(\tau)>D_k$, $h_k(x(\tau))>0$, and by continuity $\tau=t_{j+1}$.
Since every active barrier is positive, every active distance exceeds $D_k\ge D_\sigma$, and $x(t)\in\mathcal{S}_t$ for all $t\in[0,T]$.
\end{proof}

The proof uses the applied inputs only through the barrier inequalities along the realized trajectory, which the infeasible-QP counter of the experiments checks only at the sampling times.
The barrier is not continuously differentiable on a neighborhood of the closure of its zero-superlevel set, because its gradient grows without bound as $d\downarrow D_k$, so the standard invariance theorem for CBFs does not apply verbatim and the strict margin of Assumption~\ref{ass:margin} is needed.
From $h_k=0$, radial braking at exactly rate $A_k$ keeps $h_k=0$ and reaches $d_k=D_k$ in finite time.
The rule that requires the relative acceleration $A_k$ inside the margin makes the inequality discontinuous at $d_k=D_k$, which is harmless because, by the proof, the distance stays above $D_k$ while the barrier stays positive.

\subsection{Loss of safety starts with a conflict}

\begin{proof}[of Proposition~\ref{prop:conflict}]
Let $t^\circ$ be the infimum of the times $t$ at which some $k\in\K_t$ has $d_k(t)<D_k$ or $h_k(x(t))<0$, which is finite by assumption.
Since $\K_t$ changes finitely often, it is constant on $(t^\circ,t^\circ+\epsilon)$ for some $\epsilon>0$, and some constraint $k$ of this set has $d_k(t_n)<D_k$ or $h_k(x(t_n))<0$ along a sequence $t_n\downarrow t^\circ$.
Let $t_k\le t^\circ$ be the time from which $k$ is active without interruption.
If $t_k=t^\circ$, Assumption~\ref{ass:margin} and continuity give $h_k>0$ and $d_k>D_k$ near $t^\circ$, which contradicts the choice of $t_n$, so $t_k<t^\circ$.
By the definition of $t^\circ$, $d_k\ge D_k$ and $h_k\ge0$ on $[t_k,t^\circ)$.
By continuity $h_k(x(t^\circ))=0$, because either $h_k$ itself turns negative after $t^\circ$, or $d_k(t^\circ)=D_k$ with $\dot d_k(t^\circ)\le0$, and then $h_k(x(t^\circ))=n^\top\Delta v(t^\circ)=\dot d_k(t^\circ)\le0$.
Suppose that every agent of $k$ had a feasible QP for almost every $t\in[t_k,t^\circ]$.
These agents then applied inputs that satisfy their barrier inequalities, including those of $k$, for almost every $t$, and the argument in the proof of Proposition~\ref{prop:switch}, applied to $k$ alone on $[t_k,t^\circ]$, gives $h_k(x(t^\circ))\ge\eta_k(t^\circ)>0$, a contradiction.
Hence some agent $i$ of $k$ has an infeasible QP on a set $E\subseteq[t_k,t^\circ)$ of positive measure.
At every $t\in E$ all active barriers are nonnegative because $t<t^\circ$, so by part~1 of Lemma~\ref{lem:single} and $\beta_{\kappa(i)}\le\bar a_{\kappa(i)}$ each barrier inequality of $i$ is feasible on its own, and the QP of $i$ is infeasible only because its inequalities exclude each other.
\end{proof}

\subsection{Unavoidable conflicts among three robots}

\begin{proof}[of Proposition~\ref{prop:three}]
Name the robots $a$, $m$ and $c$, fix $\bar a''\in(\bar a,2\bar a)$ and $s>0$, and let $\varepsilon=s^2/(2\bar a'')$.
Robot $m$ starts at the origin at rest, $a$ at $(-(D+\varepsilon),0)$ with velocity $(s,0)$, and $c$ at $(D+\varepsilon,0)$ with velocity $(-s,0)$.
For the pairs $(a,m)$ and $(c,m)$ we have $d=D+\varepsilon$ and $n^\top\Delta v=-s$, so $h=\sqrt{4\bar a\varepsilon}-s=s\big(\sqrt{2\bar a/\bar a''}-1\big)>0$.
For the pair $(a,c)$, $h=\sqrt{4\bar a(D+2\varepsilon)}-2s>0$ whenever $s<\sqrt{\bar aD}$.
Let the three robots apply arbitrary measurable inputs with values in $[-\bar a,\bar a]^2$, write $p_j=(x_j,y_j)$, and let $t^\star=s/\bar a$.
Integrating twice gives $x_a(t^\star)\ge-(D+\varepsilon)+s^2/(2\bar a)$, $x_c(t^\star)\le D+\varepsilon-s^2/(2\bar a)$, and $|y_j(t)|\le\bar at^2/2$ for each robot, so
\begin{equation*}
x_c(t^\star)-x_a(t^\star)\le 2D-2cs^2\quad\text{with}\quad c=\tfrac12\big(1/\bar a-1/\bar a''\big)>0,
\end{equation*}
and $|y_j(t)-y_l(t)|\le s^2/\bar a$ for all robots $j,l$ and $t\le t^\star$.
If $x_m(\tau)=x_a(\tau)$ or $x_m(\tau)=x_c(\tau)$ for some $\tau\le t^\star$, the two robots concerned are at distance at most $s^2/\bar a$ at $\tau$, which is less than $D$ for $s^2<\bar aD$.
Otherwise $x_a<x_m<x_c$ on $[0,t^\star]$, the gaps $x_m-x_a$ and $x_c-x_m$ at $t^\star$ are positive and sum to at most $2D-2cs^2$, and the smaller gap is at most $D-cs^2$.
The corresponding pair is then at distance at most $\sqrt{(D-cs^2)^2+s^4/\bar a^2}$, which is less than $D$ when $s^2<2cD/(c^2+1/\bar a^2)$.
In both cases two robots are closer than $D$ at some time in $[0,t^\star]$.
\end{proof}

The construction holds for every gain $\gamma$ and every filter, because it bounds the reachable positions of all three robots directly.
At the boundary state with $h_{am}=h_{cm}=0$ and the same positions, the two barrier conditions read $u_{m,1}-u_{a,1}\ge2\bar a$ and $u_{c,1}-u_{m,1}\ge2\bar a$, which force $u_{m,1}=\bar a$ and $u_{m,1}=-\bar a$, so Definition~\ref{def:joint} fails there even for one input of the whole team.
With different bounds and capabilities, the same argument with approach speeds in the ratio $(\beta_a+\beta_m):(\beta_c+\beta_m)$ and $t^\star=(s_a+s_c)/(\bar a_a+\bar a_c)$ applies whenever $\beta_a+\beta_c+2\beta_m>\bar a_a+\bar a_c$, which always holds for $\beta=\bar a$.

\paragraph{Chains of robots.}
Let $m$ robots with bound $\bar a$ and capability $\beta=\theta\bar a$ start on the first axis at $x_j(0)=(j-1)(D+\varepsilon)$ with velocities $v_j(0)=\big(\tfrac{m+1}{2}-j\big)\sigma\,e_1$, so that every adjacent gap closes at speed $\sigma$, and let $\varepsilon=\sigma^2/(4\theta''\bar a)$ with $\theta''\in(1/(m-1),\theta)$.
Every adjacent barrier, with $A_k=2\theta\bar a$, equals $\sigma(\sqrt{\theta/\theta''}-1)>0$, and the other barriers are positive for small $\sigma$.
The inputs of the inner robots do not affect $x_m-x_1$, and braking the two outer robots as hard as possible gives, at $t^\star=(m-1)\sigma/(2\bar a)$,
\begin{equation*}
x_m(t^\star)-x_1(t^\star)\le(m-1)(D+\varepsilon)-\frac{(m-1)^2\sigma^2}{4\bar a}=(m-1)D-\frac{(m-1)\sigma^2}{4\bar a}\Big((m-1)-\frac{1}{\theta''}\Big),
\end{equation*}
which is less than $(m-1)D$ because $\theta''(m-1)>1$.
The crossing and lateral bounds of the three-robot case then give two robots closer than $D$ for small $\sigma$.
At the state with the same positions, $\varepsilon\le D/(m-1)$ and every adjacent barrier zero, all other barriers are nonnegative, and the adjacent barrier conditions $u_{j+1,1}-u_{j,1}\ge2\theta\bar a$ add up to $u_{m,1}-u_{1,1}\ge2(m-1)\theta\bar a>2\bar a$, which no inputs in $[-\bar a,\bar a]^2$ achieve, so Definition~\ref{def:joint} fails there.
Hence, for every $\theta\in(0,1]$, a chain of $m>1+1/\theta$ robots contains states with positive barriers from which a collision is unavoidable, and no fixed reduction of the capabilities makes pairwise braking barriers jointly valid on their whole safe set for all team sizes.

\section{Ontology and reasoning}\label{app:dl}

\begin{table}[h]
\centering\small
\caption{Class profiles with the per-axis acceleration limit $\bar a$ in m/s$^2$ and the nominal speed $\bar v$ of the task controller in m/s. Robot--human constraints need $\beta_C>\bar a_h$, which holds for every robot class.}
\label{tab:profiles}
\begin{tabular}{lllcc}
\toprule
Class & Parent & Role in the experiments & $\bar a$ & $\bar v$\\
\midrule
\cls{Drone} & \cls{Aerial} & design class & 2.0 & 1.5\\
\cls{LightDrone} & \cls{Drone} & test only, $\dyn\cls{Drone}$, inherits & 3.0 & 1.5\\
\cls{HeavyDrone} & \cls{Drone} & test only, $\ndyn\cls{Drone}$, own capability & 1.0 & 1.5\\
\cls{Rover} & \cls{Ground} & design class & 1.2 & 1.0\\
\cls{Carrier} & \cls{Ground} & design class & 1.0 & 0.8\\
\cls{Forklift} & \cls{Ground} & test only, no design ancestor & 1.2 & 1.0\\
\cls{Human} & \cls{UncontrolledAgent} & uncontrolled & 0.5 (disk) & 0.8\\
\cls{Payload} & \cls{Object} & design class & -- & --\\
\cls{HazardousPayload} & \cls{Payload} & design class & -- & --\\
\cls{Chemical} & \cls{HazardousPayload} & test only & -- & --\\
\bottomrule
\end{tabular}
\end{table}

{\sloppy Besides the taxonomy of Table~\ref{tab:profiles}, the TBox contains $\cls{ControlledAgent}\sqsubseteq\cls{Agent}$, $\cls{UncontrolledAgent}\sqsubseteq\cls{Agent}$, $\cls{Aerial},\cls{Ground}\sqsubseteq\cls{ControlledAgent}$, $\cls{Payload},\cls{RestrictedZone}\sqsubseteq\cls{Object}$, $\cls{Agent},\cls{Object}\sqsubseteq\cls{Entity}$, the disjointness axioms for $\cls{Agent}\sqcap\cls{Object}$, $\cls{ControlledAgent}\sqcap\cls{UncontrolledAgent}$, $\cls{Aerial}\sqcap\cls{Ground}$ and $\cls{Payload}\sqcap\cls{RestrictedZone}$, the symmetric role $\mathrm{near}$ on \cls{Entity}, the role $\mathrm{carries}$ from \cls{ControlledAgent} to \cls{Payload}, the role $\mathrm{nearCarried}$ from \cls{ControlledAgent} to \cls{Entity}, the chain $\mathrm{carries}\circ\mathrm{near}\sqsubseteq\mathrm{nearCarried}$, and the definition $\cls{HazardCarrier}\equiv\cls{ControlledAgent}\sqcap\exists\mathrm{carries}.\cls{HazardousPayload}$.
The safety axioms are $\mathrm{near}(\cls{ControlledAgent},\cls{ControlledAgent})$ with 0.3\,m, $\mathrm{near}(\cls{ControlledAgent},\cls{Human})$ with 0.5\,m, $\mathrm{near}(\cls{Aerial},\cls{Human})$ with 1.0\,m, $\mathrm{near}(\cls{Ground},\cls{RestrictedZone})$ with 1.0\,m from the zone center, $\mathrm{nearCarried}(\cls{ControlledAgent},\cls{Human})$ with 0.8\,m, and $\mathrm{nearCarried}(\cls{HazardCarrier},\cls{Human})$ with 2.0\,m, the last two measured from the payload.\par}
The OWL file used in all experiments is released with the code.

\paragraph{Completeness of the compiled reasoner.}
Apart from the definition of \cls{HazardCarrier}, every TBox axiom above lies in OWL~2~RL \citep{owl2profiles}, namely the inclusions and disjointness axioms between concept names, the domain and range restrictions, the symmetric role and the property chain.
The definition contributes the inclusions $\cls{HazardCarrier}\sqsubseteq\cls{ControlledAgent}$ and $\cls{ControlledAgent}\sqcap\exists\mathrm{carries}.\cls{HazardousPayload}\sqsubseteq\cls{HazardCarrier}$, which are RL rules, and the inclusion $\cls{HazardCarrier}\sqsubseteq\exists\mathrm{carries}.\cls{HazardousPayload}$, whose existential right-hand side lies outside OWL~2~RL, so we argue directly.
Let $\mathcal{M}$ be the least fixpoint of the ABox under the rules $A(x)\to B(x)$ for atomic inclusions, $r(x,y)\to\mathrm{Dom}_r(x)$ and $r(x,y)\to\mathrm{Ran}_r(y)$, symmetry of $\mathrm{near}$, the chain $\mathrm{carries}(x,y)\wedge\mathrm{near}(y,z)\to\mathrm{nearCarried}(x,z)$, the two RL inclusions of the definition, and $A(x)\wedge B(x)\to\bot$ for each disjointness axiom.
If $\bot\notin\mathcal{M}$, extending $\mathcal{M}$ by one fresh \cls{HazardousPayload} carried by every \cls{HazardCarrier} that carries none gives a model of $\T\cup\A$, because the fresh payload has no $\mathrm{near}$ relation and its classes \cls{HazardousPayload}, \cls{Payload}, \cls{Object} and \cls{Entity} cause no clash.
Every rule is sound, so $\T\cup\A$ is inconsistent if and only if $\bot\in\mathcal{M}$, and otherwise it entails a class or role assertion about named individuals if and only if the assertion is in $\mathcal{M}$.
The argument relies on $\mathrm{carries}$ having no functionality or cardinality restriction.
The compiled reasoner computes $\mathcal{M}$, and its agreement with HermiT on \res{testTrials} random ABoxes with 5 to 9 individuals, some of them inconsistent, checks the implementation.

\paragraph{Inconsistency.}
If $\T\cup\A_t$ is inconsistent, it has no model and entails every assertion, in particular $\mathrm{near}(a,a)$ and $\cls{ControlledAgent}(a)$ for every agent $a$.
The instance $(\sigma,a,a)$ of the robot--robot axiom is then active and requires $\|p_a-p_a\|\ge0.3$\,m, which every state violates, so $\mathcal{S}_t=\emptyset$.
In a consistent ABox every active constraint relates two distinct entities, because \eqref{eq:abox} asserts $\mathrm{near}$ only between distinct entities and the self-pair $\mathrm{nearCarried}(a,a)$, derived from a carried payload near its carrier, matches only axioms whose second class is \cls{Human}, which is disjoint from \cls{ControlledAgent}.
The compiled consistency check tests whether $\bot\in\mathcal{M}$, so it also finds clashes that the taxonomy propagates, such as the one below.
In each of \res{incN} W0 episodes, with seeds disjoint from the evaluation seeds, a faulty transponder makes the worker closest to a robot also report the class \cls{Rover} for 3\,s, so that one individual is typed both \cls{Human} and \cls{Rover}, which the TBox makes disjoint through $\cls{ControlledAgent}\sqcap\cls{UncontrolledAgent}\sqsubseteq\bot$.
A filter that trusts the label treats the worker as a robot, with a 0.3\,m keep-out and no worst-case human acceleration, and within the fault and the 2\,s after it \res{incTrustEps} episodes violate safety (\res{incTrustSteps} violating steps and \res{incTrustCf} single-constraint failures).
The compiled check detects \res{incDet} of the \res{incN} faults at the event, and HermiT agrees on \res{incHermit} sampled events.
OntoCBF then keeps the last consistent ABox and brakes the team until the contradiction disappears, after which \res{incOntoEps} episodes violate safety (\res{incOntoSteps} steps), at the cost of \res{incDelivOnto} against \res{incDelivRef} deliveries per episode on the same seeds without a fault.

\paragraph{Labels and defaults.}
Entailment is monotone, so an entity whose class label is missing or too general receives only the constraints of the classes entailed for it, and an entity typed only as \cls{Entity} receives none.
Class labels and dynamics profiles are therefore safety-relevant inputs, and the profiles must be measured under the loads a robot carries.
A fail-safe default, such as typing every sensed entity of unknown class as \cls{Human}, gives such an entity the largest keep-out distances among uncontrolled entities and the worst-case human acceleration.
This default belongs to the layer that builds the ABox, because a description logic draws conclusions only from the axioms and assertions that are present.

\section{Experimental details and additional results}\label{app:exp}

\begin{figure}[h]
\centering
\includegraphics[width=\linewidth]{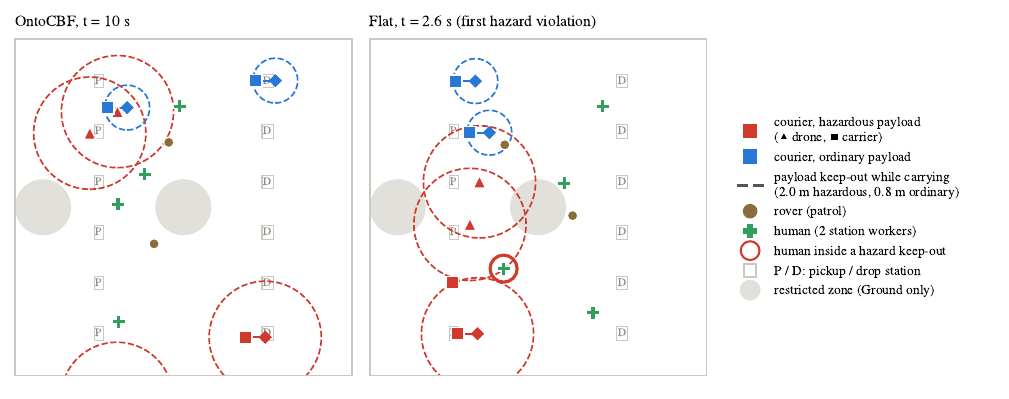}
\caption{Scenario W3 (a \cls{LightDrone} carries a \cls{Chemical} and a \cls{Drone} a \cls{HazardousPayload}), same seed under OntoCBF ($t=10$\,s) and under Flat at its first hazard violation ($t=2.6$\,s). For a drone that carries a hazardous payload, Flat falls back to its default payload keep-out and keeps the payload only 0.8\,m from people.}
\label{fig:warehouse}
\end{figure}

\paragraph{Warehouse layout.}
Each $12\times12$\,m tile has six lanes from a pickup station at $x=3$\,m to a drop station at $x=9$\,m (Fig.~\ref{fig:warehouse}).
The two middle lanes belong to the drones and cross a restricted zone at the center of the tile, which binds only \cls{Ground} agents, and the four outer lanes belong to the carriers.
A second zone lies west of the pickup column.
The payload of a drone hangs below it, and the payload of a carrier sits 0.7\,m from it along $+x$, toward the drop station.
After a drop, the next item waits at the pickup station.
Two of the four humans per tile are station workers who walk between the pickup and drop areas of their tile, and the other humans and the rovers move between random points of the whole floor.
W4 tiles this layout $2\times2$ on a $24\times24$\,m floor.

\begin{figure}[h]
\centering
\includegraphics[width=0.92\linewidth]{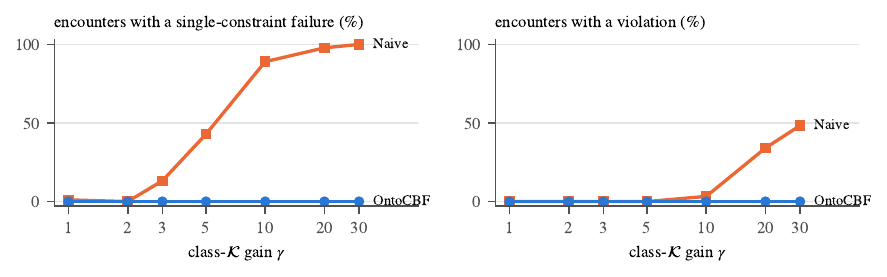}
\caption{Stress test of the capability assignment for \cls{HeavyDrone} (Section~\ref{sec:capability}) in head-on encounters with a human, all starting inside the safe set of the true capability. Under Naive the \cls{HeavyDrone} inherits the capability of \cls{Drone}, and OntoCBF assigns it its own capability. The left panel shows the encounters with a single-constraint failure and the right panel those with a violation.}
\label{fig:stress}
\end{figure}

\paragraph{Gain selection.}
On 100 W0 validation episodes, with seeds disjoint from the evaluation seeds, OntoCBF with the PD controller violates safety in \res{gselList}\,\% of the episodes for $\gamma=\res{gselGammas}$, with \res{gselDelLo} to \res{gselDelHi} deliveries per courier.
The selection rule was fixed before evaluation and takes the largest $\gamma$ whose violation rate is at most the rate at $\gamma=1$.
Since the rate grows with $\gamma$, the rule selects $\gamma=\res{gammaMain}$, the most conservative gain of the grid, which is also the gain at which an overestimated capability is least visible (Fig.~\ref{fig:stress} and Table~\ref{tab:gamma}).

\paragraph{Stress test of the capability assignment (Section~\ref{sec:capability}).}
A human walks straight at a robot whose goal lies behind the human, with initial distances of 2 to 6\,m, robot speeds of 0.5 to 1.5\,m/s and human speeds of 0 to 0.8\,m/s.
Only initial states in the safe set of the aerial--human barrier with the true capability of the class are kept (\res{stN} encounters per setting for \cls{HeavyDrone}), so with this capability every single constraint stays feasible by Lemma~\ref{lem:single}.
We sweep the gain $\gamma$ of $\dot h\ge-\gamma h^3$, which is \res{gammaMain} in the warehouse (Fig.~\ref{fig:stress}).
OntoCBF avoids single-constraint failures and violations at all gains, with a minimum distance of \res{stOntoDmin}\,m, and for \cls{Drone}, whose capability is the same under both methods, Naive and OntoCBF coincide.
At $\gamma=\res{gammaMain}$, single constraints fail in \res{stNaiveCfMain}\,\% of the encounters under Naive, and the \cls{HeavyDrone} stays outside the keep-out region in all of them, so in the warehouse an overestimated capability shows up mainly as single-constraint failures.

\paragraph{The dynamic-subtype test in the warehouse at larger gains.}
Table~\ref{tab:gamma} repeats W1, the only scenario with a class that fails the test, for Naive and OntoCBF at larger $\gamma$, and its row for $\gamma=1$ reproduces the main results seed by seed.
OntoCBF violates safety in \res{gnOntoLo}--\res{gnOntoHi}\,\% of the episodes at every gain.
Naive violates safety in \res{gn/1/vpN}\,\% of the episodes at $\gamma=1$ (McNemar $p={}$\res{gn/1/p} against OntoCBF on the same seeds), in \res{gn/3/vpN}\,\% at $\gamma=3$, where its aerial--human violations are significantly more frequent, and in \res{gn/10/vpN}\,\% at $\gamma=10$ ($p$\,\res{gn/10/p}).
Its overestimated capability lets the \cls{HeavyDrone} brake later, which also yields slightly more deliveries.

\begin{table}[h]
\centering\small
\caption{Naive and OntoCBF in W1 (unseen robots) as the class-$\mathcal{K}$ gain grows, with the PD controller and \res{nSeeds} paired seeds ($\gamma=1$ is the main setting). Differences carry paired 95\,\% bootstrap intervals, and $p$ comes from McNemar's exact test on episodes with a violation. OntoCBF has zero single-constraint failures at every gain.}
\label{tab:gamma}
\resizebox{\linewidth}{!}{\begin{tabular}{r rr r rr r rr}
\toprule
 & \multicolumn{2}{c}{Episodes with a violation (\%)} & & \multicolumn{2}{c}{Naive $-$ OntoCBF, per episode} & Single-constraint & \multicolumn{2}{c}{Deliveries / courier} \\
\cmidrule(lr){2-3}\cmidrule(lr){5-6}\cmidrule(lr){8-9}
$\gamma$ & Naive & OntoCBF & McNemar $p$ & violating steps & aerial--human & failures (Naive) & Naive & OntoCBF \\
\midrule
1 & 23 & 20 & 0.63 & $+0.8$ [$-2.6$, $+4.3$] & $-0.2$ [$-1.5$, $+1.0$] & 266 & 1.79 & 1.64 \\
3 & 27 & 21 & 0.26 & $+1.9$ [$-0.6$, $+4.3$] & $+3.1$ [$+0.4$, $+6.1$] & 907 & 2.00 & 1.87 \\
10 & 44 & 21 & $<$0.001 & $+2.9$ [$-0.6$, $+6.3$] & $+2.3$ [$-3.1$, $+7.3$] & 2223 & 2.13 & 2.02 \\
\bottomrule
\end{tabular}
}
\end{table}

\begin{table}[h]
\centering\small
\caption{Episodes with a violation (\%) in the warehouse with 95\,\% Wilson intervals, PD controller, \res{nSeeds} seeds per cell.}
\label{tab:ci}
\resizebox{\linewidth}{!}{\begin{tabular}{lcccccccc}
\toprule
Scenario & None & Flat & Flat-cons & No-chain & Naive & Untyped & No-act. & OntoCBF \\
\midrule
W0 (design) & 100 [96, 100] & 20 [13, 29] & 20 [13, 29] & 100 [96, 100] & 22 [15, 31] & 16 [10, 24] & 74 [65, 82] & 22 [15, 31] \\
W1 (unseen robots) & 100 [96, 100] & 89 [81, 94] & 17 [11, 26] & 100 [96, 100] & 23 [16, 32] & 17 [11, 26] & 72 [63, 80] & 20 [13, 29] \\
W2 (unseen payload) & 100 [96, 100] & 99 [95, 100] & 20 [13, 29] & 100 [96, 100] & 22 [15, 31] & 16 [10, 24] & 74 [65, 82] & 22 [15, 31] \\
W3 (unseen combination) & 100 [96, 100] & 100 [96, 100] & 28 [20, 37] & 100 [96, 100] & 40 [31, 50] & 37 [28, 47] & 96 [90, 98] & 40 [31, 50] \\
W4 ($N{=}32$) & 100 [96, 100] & 67 [57, 75] & 67 [57, 75] & 100 [96, 100] & 70 [60, 78] & 69 [59, 77] & 98 [93, 99] & 70 [60, 78] \\
\bottomrule
\end{tabular}
}
\end{table}

\begin{sidewaystable}
\centering\small
\caption{Violations by axiom in the warehouse (constraint-steps per episode), deliveries per courier for ordinary and hazardous payloads, and pickups that began inside a keep-out region (out of all pickups). The MAPPO entries pool \res{nRuns}$\times$\res{nSeeds} episodes.}
\label{tab:axioms}
\resizebox{\linewidth}{!}{\begin{tabular}{l l rrrrrr rr r}
\toprule
Scenario & Method & hazard--human 2.0\,m & payload--human 0.8\,m & aerial--human 1.0\,m & robot--human 0.5\,m & ground--zone 1.0\,m & robot--robot 0.3\,m & \multicolumn{2}{c}{Deliveries/courier} & Pickups into viol. \\
 & &  &  &  &  &  &  & ordinary & hazardous & \\
\midrule
W0 (design) & None & 423.0 & 252.2 & 344.3 & 250.9 & 71.9 & 63.7 & 3.00 & 2.00 & 348/2200 \\
W0 (design) & Flat & 5.4 & 1.2 & 2.3 & 0.3 & 1.2 & 0.0 & 2.05 & 1.16 & 0/1373 \\
W0 (design) & Flat-cons & 5.4 & 1.2 & 2.3 & 0.3 & 1.2 & 0.0 & 2.05 & 1.16 & 0/1373 \\
W0 (design) & No-chain & 421.5 & 78.6 & 3.2 & 0.3 & 1.1 & 0.0 & 2.12 & 1.65 & 126/1500 \\
W0 (design) & Naive & 4.4 & 0.8 & 2.2 & 0.2 & 1.3 & 0.0 & 2.06 & 1.16 & 0/1373 \\
W0 (design) & Untyped & 5.7 & 0.1 & 0.6 & 0.1 & 1.0 & 0.0 & 1.36 & 1.18 & 0/1136 \\
W0 (design) & No-act. & 20.2 & 1.5 & 2.2 & 0.2 & 1.1 & 0.0 & 2.06 & 1.18 & 99/1391 \\
W0 (design) & OntoCBF & 4.4 & 0.8 & 2.2 & 0.2 & 1.3 & 0.0 & 2.06 & 1.16 & 0/1373 \\
W0 (design) & MAPPO & 0.9 & 0.9 & 1.1 & 0.6 & 1.4 & 0.1 & 1.95 & 1.28 & 0/4042 \\
W1 (unseen robots) & None & 423.0 & 247.2 & 336.2 & 244.9 & 71.9 & 64.9 & 3.25 & 2.00 & 326/2100 \\
W1 (unseen robots) & Flat & 3.3 & 2.4 & 76.2 & 1.4 & 1.4 & 0.0 & 1.95 & 1.17 & 0/1343 \\
W1 (unseen robots) & Flat-cons & 4.8 & 0.3 & 0.8 & 0.5 & 0.6 & 0.0 & 1.00 & 1.18 & 0/996 \\
W1 (unseen robots) & No-chain & 412.9 & 103.9 & 1.9 & 0.6 & 0.9 & 0.0 & 1.92 & 1.66 & 139/1433 \\
W1 (unseen robots) & Naive & 4.1 & 0.5 & 1.8 & 0.8 & 2.2 & 0.0 & 2.10 & 1.17 & 0/1402 \\
W1 (unseen robots) & Untyped & 5.0 & 0.0 & 0.7 & 0.5 & 1.0 & 0.0 & 1.23 & 1.19 & 0/1096 \\
W1 (unseen robots) & No-act. & 20.8 & 1.8 & 1.9 & 0.2 & 0.6 & 0.0 & 1.87 & 1.20 & 102/1331 \\
W1 (unseen robots) & OntoCBF & 3.1 & 0.5 & 1.9 & 0.2 & 1.0 & 0.0 & 1.87 & 1.19 & 0/1315 \\
W1 (unseen robots) & MAPPO & 0.9 & 0.7 & 2.2 & 0.7 & 1.0 & 0.0 & 1.89 & 1.29 & 0/3976 \\
W2 (unseen payload) & None & 423.0 & 252.2 & 344.3 & 250.9 & 71.9 & 63.7 & 3.00 & 2.00 & 348/2200 \\
W2 (unseen payload) & Flat & 359.7 & 2.0 & 2.2 & 0.0 & 0.9 & 0.0 & 2.06 & 1.56 & 115/1451 \\
W2 (unseen payload) & Flat-cons & 5.4 & 1.2 & 2.3 & 0.3 & 1.2 & 0.0 & 2.05 & 1.16 & 0/1373 \\
W2 (unseen payload) & No-chain & 421.5 & 78.6 & 3.2 & 0.3 & 1.1 & 0.0 & 2.12 & 1.65 & 126/1500 \\
W2 (unseen payload) & Naive & 4.4 & 0.8 & 2.2 & 0.2 & 1.3 & 0.0 & 2.06 & 1.16 & 0/1373 \\
W2 (unseen payload) & Untyped & 5.7 & 0.1 & 0.6 & 0.1 & 1.0 & 0.0 & 1.36 & 1.18 & 0/1136 \\
W2 (unseen payload) & No-act. & 20.2 & 1.5 & 2.2 & 0.2 & 1.1 & 0.0 & 2.06 & 1.18 & 99/1391 \\
W2 (unseen payload) & OntoCBF & 4.4 & 0.8 & 2.2 & 0.2 & 1.3 & 0.0 & 2.06 & 1.16 & 0/1373 \\
W2 (unseen payload) & MAPPO & 0.9 & 0.9 & 1.1 & 0.6 & 1.4 & 0.1 & 1.95 & 1.28 & 0/4042 \\
W3 (unseen combination) & None & 1052.7 & 255.4 & 350.1 & 252.3 & 71.9 & 62.2 & 2.00 & 3.00 & 674/2200 \\
W3 (unseen combination) & Flat & 840.6 & 0.5 & 41.8 & 0.2 & 0.9 & 0.0 & 1.62 & 1.71 & 156/1342 \\
W3 (unseen combination) & Flat-cons & 17.8 & 0.5 & 0.4 & 0.3 & 1.3 & 0.0 & 1.65 & 1.12 & 0/1112 \\
W3 (unseen combination) & No-chain & 1237.7 & 82.0 & 2.8 & 0.3 & 1.1 & 0.0 & 1.75 & 2.09 & 281/1511 \\
W3 (unseen combination) & Naive & 28.2 & 0.1 & 0.5 & 0.2 & 1.8 & 0.1 & 1.60 & 1.34 & 0/1197 \\
W3 (unseen combination) & Untyped & 27.3 & 0.1 & 0.8 & 0.2 & 1.1 & 0.1 & 1.28 & 1.33 & 0/1140 \\
W3 (unseen combination) & No-act. & 60.4 & 0.7 & 0.2 & 0.2 & 1.4 & 0.1 & 1.59 & 1.36 & 217/1237 \\
W3 (unseen combination) & OntoCBF & 28.2 & 0.1 & 0.5 & 0.2 & 1.8 & 0.1 & 1.60 & 1.34 & 0/1197 \\
W3 (unseen combination) & MAPPO & 13.3 & 0.2 & 0.5 & 0.4 & 1.3 & 0.0 & 1.69 & 1.29 & 0/3529 \\
W4 ($N{=}32$) & None & 1561.8 & 874.6 & 1111.5 & 846.3 & 239.2 & 191.0 & 3.00 & 2.00 & 1244/8800 \\
W4 ($N{=}32$) & Flat & 47.8 & 6.1 & 8.3 & 3.8 & 10.8 & 0.0 & 2.12 & 1.16 & 0/5667 \\
W4 ($N{=}32$) & Flat-cons & 47.8 & 6.1 & 8.3 & 3.8 & 10.8 & 0.0 & 2.12 & 1.16 & 0/5667 \\
W4 ($N{=}32$) & No-chain & 1588.0 & 297.9 & 9.2 & 3.8 & 2.3 & 0.0 & 2.19 & 1.66 & 369/6092 \\
W4 ($N{=}32$) & Naive & 55.5 & 6.5 & 9.6 & 3.6 & 10.8 & 0.0 & 2.13 & 1.16 & 0/5666 \\
W4 ($N{=}32$) & Untyped & 55.6 & 4.0 & 5.4 & 3.8 & 13.7 & 0.3 & 1.48 & 1.14 & 0/4768 \\
W4 ($N{=}32$) & No-act. & 122.2 & 10.1 & 9.2 & 4.3 & 13.1 & 0.0 & 2.12 & 1.16 & 296/5686 \\
W4 ($N{=}32$) & OntoCBF & 55.5 & 6.5 & 9.6 & 3.6 & 10.8 & 0.0 & 2.13 & 1.16 & 0/5666 \\
W4 ($N{=}32$) & MAPPO & 44.1 & 5.1 & 7.1 & 4.0 & 7.8 & 0.1 & 1.98 & 1.10 & 0/15980 \\
\bottomrule
\end{tabular}
}
\end{sidewaystable}

\paragraph{Single-element ablations.}
Each ablation changes exactly one element of OntoCBF (Table~\ref{tab:ingredients}), and Table~\ref{tab:single} shows in which scenarios the three ablations below fail.

\begin{table}[h]
\centering\small
\caption{Ablations that remove one element of OntoCBF, each in the scenario that needs the element. Violations are counted in constraint-steps per episode, and single-constraint failures and pickups into a violation are summed over \res{nSeeds} episodes.}
\label{tab:ingredients}
\resizebox{0.8\linewidth}{!}{\begin{tabular}{l l l rr}
\toprule
Scenario & Without & Measure & Ablation & OntoCBF \\
\midrule
W0 (design) & property chain & payload+hazard viol. & 500.0 & 5.2 \\
W1 (unseen robots) & robot taxonomy & aerial--human viol. & 150.7 & 1.9 \\
W1 (unseen robots) & dynamic-subtype test & single-constraint failures (total) & 266 & 0 \\
W2 (unseen payload) & payload taxonomy & hazard--human viol. & 360.1 & 4.4 \\
W3 (unseen combination) & defined class & hazard--human viol. & 835.0 & 28.2 \\
W3 (unseen combination) & activation check & pickups into a violation (total) & 217 & 0 \\
W0 (design) & hazard distinction & deliveries / courier (ordinary) & 1.36 & 2.06 \\
\bottomrule
\end{tabular}
}
\end{table}

\emph{No robot tax.}\ types every robot outside the design classes as \cls{ControlledAgent} only, so all its ancestors lie above the design classes and it is assigned its own profile.
\emph{No payload tax.}\ types every payload class outside the design classes as \cls{Payload}.
\emph{No defined class} skips the realization of \cls{HazardCarrier} and writes the hazard keep-out only for the design pair (\cls{Carrier}, \cls{HazardousPayload}), closed under both taxonomies.

\begin{table}[h]
\centering\small
\caption{Episodes with a violation (\%) under single-element ablations where the element is needed, PD controller, \res{nSeeds} seeds. The sign $=$ marks scenarios in which the ablation acts exactly as OntoCBF, seed by seed, and all three ablations have zero single-constraint failures.}
\label{tab:single}
\resizebox{0.75\linewidth}{!}{\begin{tabular}{l rrrr}
\toprule
Scenario & No robot tax. & No payload tax. & No defined class & OntoCBF \\
\midrule
W0 (design) & $=$ & $=$ & $=$ & 22 \\
W1 (unseen robots) & 100 & $=$ & $=$ & 20 \\
W2 (unseen payload) & $=$ & 99 & $=$ & 22 \\
W3 (unseen combination) & 79 & 100 & 100 & 40 \\
W4 ($N{=}32$) & $=$ & $=$ & $=$ & 70 \\
\bottomrule
\end{tabular}
}
\end{table}

\begin{figure}[h]
\centering
\includegraphics[width=\linewidth]{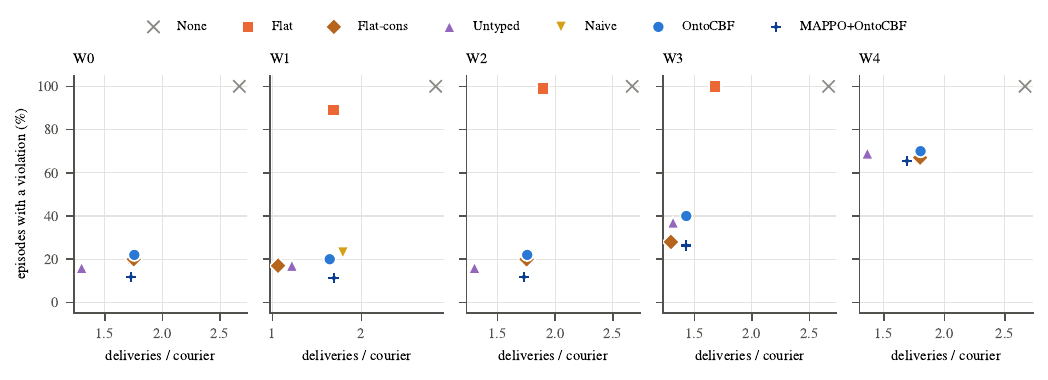}
\caption{Episodes with a violation against deliveries per courier, \res{nSeeds} seeds per point and \res{nRuns} pooled training runs for MAPPO. Points further down and to the right are better.}
\label{fig:pareto}
\end{figure}

\paragraph{Specification effort and throughput.}
W3 alone accounts for \res{effOntoW3} of the \res{effOnto} ontology statements of OntoCBF and \res{effFlatW3} of the \res{effFlat} table entries of Flat that cover W1--W3.
Fig.~\ref{fig:pareto} places each method by deliveries per courier and episodes with a violation.

\begin{table}[h]
\centering\small
\caption{Deliveries per courier in the warehouse without a filter and with OntoCBF, and computational cost of OntoCBF with the PD controller on one CPU core.}
\label{tab:cost}
\resizebox{\linewidth}{!}{\begin{tabular}{l rr r rrr}
\toprule
Scenario & \multicolumn{2}{c}{Deliveries/courier} & QP infeas.\ (\%) & Filter (ms/step) & Events/ep. & Reasoning (ms/event) \\
 & None & OntoCBF & & & & \\
\midrule
W0 (design) & 2.67 & 1.76 & 1.2 & 1.40 & 268 & 0.91 \\
W1 (unseen robots) & 2.83 & 1.64 & 1.1 & 1.42 & 267 & 0.91 \\
W2 (unseen payload) & 2.67 & 1.76 & 1.2 & 1.40 & 268 & 0.91 \\
W3 (unseen combination) & 2.67 & 1.43 & 1.6 & 1.58 & 262 & 0.93 \\
W4 ($N{=}32$) & 2.67 & 1.80 & 2.2 & 6.10 & 774 & 4.60 \\
\bottomrule
\end{tabular}
}
\end{table}

\paragraph{Events before a violation.}
For every violation onset of OntoCBF with PD in W0, W1, W3 and W4 (W2 replays W0), we identify the last infeasible QP of an agent of the violated constraint.
An infeasible QP is a conflict if each of its inequalities is feasible on its own in the true input set.
As a base rate, we give the share of agent-steps whose agent had a conflict within the window, over all agent-steps and over those in a close encounter, meaning an active constraint within 1\,m of its keep-out.
Conflicts make up \res{vcJointShare}\,\% of all infeasible QPs, and in the others every inequality that is infeasible on its own has an already negative barrier, as Proposition~\ref{prop:conflict} predicts.
We call an onset robot-attributable if, at the onset, a robot of the violated constraint has a positive velocity component toward the other entity.
Section~\ref{sec:cost} reports the complementary share, the onsets at which the robot was not moving toward the other entity.

\begin{table}[h]
\centering\small
\caption{Violation onsets of OntoCBF with PD preceded by a conflict of an agent of the violated constraint, and base rates over agent-steps (window 1\,s).}
\label{tab:cause}
\resizebox{0.85\linewidth}{!}{\begin{tabular}{l r rr rr}
\toprule
 & & \multicolumn{2}{c}{Onsets after a conflict} & \multicolumn{2}{c}{Base rate (agent-steps)} \\
\cmidrule(lr){3-4}\cmidrule(lr){5-6}
Scenario & Onsets & within 1\,s & within 2\,s & all & close encounter \\
\midrule
W0 (design) & 45 & 45 & 45 & 2.0\,\% & 3.9\,\% \\
W1 (unseen robots) & 31 & 31 & 31 & 1.9\,\% & 3.7\,\% \\
W3 (unseen combination) & 108 & 108 & 108 & 2.6\,\% & 4.6\,\% \\
W4 ($N{=}32$) & 282 & 282 & 282 & 3.3\,\% & 6.4\,\% \\
\bottomrule
\end{tabular}
}
\end{table}

\paragraph{Joint validity on visited states.}
On every step of the \res{nSeeds} W0 evaluation episodes under OntoCBF with PD, we take the active family and ask whether one input for the whole team, inside every agent's true input set, satisfies all barrier conditions at once, with robot pairs unsplit and worst-case human accelerations.
This is a linear feasibility problem, solved with HiGHS.
We count only states at which every active barrier of the configuration has $h_k\ge0$ outside the margin.
With \cls{LightDrone}s the inequalities are those of the design team and the input sets are larger, so validity can only grow by Theorem~\ref{thm:inherit}, and this configuration tests the implementation.
With \cls{HeavyDrone}s under the capability of \cls{Drone}, the inequalities are the same and the input sets are smaller.
\cls{HeavyDrone}s with their own capability form a different family, which is valid at \res{jvOwnPct}\,\% of the visited states at which the design team's family is valid.
Feasibility of every agent's QP is lower than joint validity throughout, because the responsibility split is more restrictive than a joint input.

\begin{table}[h]
\centering\small
\caption{Joint validity (Definition~\ref{def:joint}) on the states visited by the design team in W0, with both drones substituted.}
\label{tab:joint}
\resizebox{\linewidth}{!}{\begin{tabular}{l rrrr}
\toprule
Configuration (two drones substituted) & States & Jointly valid & All per-agent QPs feasible & Valid where the design team is \\
\midrule
\cls{Drone} (design) & 78837 & 94.43\,\% & 92.44\,\% & 74442/74442 \\
\cls{LightDrone}, inherited $\beta{=}2.0$ & 78837 & 94.57\,\% & 92.56\,\% & 74442/74442 \\
\cls{HeavyDrone}, $\beta{=}2.0$ (Naive) & 78837 & 93.34\,\% & 91.56\,\% & 73589/74442 \\
\cls{HeavyDrone}, own $\beta{=}1.0$ & 64845 & 87.11\,\% & 84.79\,\% & 56485/61668 \\
\bottomrule
\end{tabular}
}
\end{table}

\begin{table}[h]
\centering\small
\caption{PD and MAPPO behind OntoCBF, \res{nSeeds} paired seeds per cell. MAPPO pools \res{nRuns} runs trained in W0 only, so its cells hold \res{nRuns}$\times$\res{nSeeds} episodes, and the intervals in Section~\ref{sec:experiments} resample training runs and then seeds. The columns give violating steps per episode (Viol.), infeasible QPs as a percentage of agent-steps with active constraints (Inf.) and deliveries per courier (Deliv.), and the last column gives the percentage of episodes with a violation for MAPPO without the filter.}
\label{tab:learned}
\resizebox{0.8\linewidth}{!}{\begin{tabular}{l rrr rrr r}
\toprule
 & \multicolumn{3}{c}{PD} & \multicolumn{3}{c}{MAPPO} & No filter \\
\cmidrule(lr){2-4}\cmidrule(lr){5-7}\cmidrule(lr){8-8}
Scenario & Viol. & Inf.\,\% & Deliv. & Viol. & Inf.\,\% & Deliv. & Viol.\ ep.\,\% \\
\midrule
W0 (design) & 6.1 & 1.20 & 1.76 & 3.1 & 0.73 & 1.73 & 100 \\
W1 (unseen robots) & 5.6 & 1.15 & 1.64 & 3.3 & 0.75 & 1.69 & 100 \\
W2 (unseen payload) & 6.1 & 1.20 & 1.76 & 3.1 & 0.73 & 1.73 & 100 \\
W3 (unseen combination) & 16.0 & 1.62 & 1.43 & 9.0 & 0.95 & 1.43 & 100 \\
W4 ($N{=}32$) & 47.9 & 2.21 & 1.80 & 39.6 & 1.81 & 1.69 & 100 \\
\bottomrule
\end{tabular}
}
\end{table}

\begin{table}[h]
\centering\small
\caption{MAPPO minus PD behind OntoCBF for each training run (W2 equals W0), with paired 95\,\% bootstrap intervals over the \res{nSeeds} seeds (\res{nBoot} resamples). Violating steps fall significantly in \res{nViolDownRuns} of the \res{nRunCells} run--scenario cells.}
\label{tab:runs}
\resizebox{0.9\linewidth}{!}{\begin{tabular}{l l rrr}
\toprule
Scenario & Run & $\Delta$ infeasible QPs / ep. & $\Delta$ violating steps / ep. & $\Delta$ deliveries / courier \\
\midrule
W0 (design) & 1 & $-23.9$ [$-37.7$, $-10.5$] & $-1.3$ [$-5.3$, $+2.7$] & $-0.01$ [$-0.05$, $+0.04$] \\
 & 2 & $-34.4$ [$-47.6$, $-21.6$] & $-4.4$ [$-7.8$, $-1.6$] & $+0.04$ [$-0.01$, $+0.08$] \\
 & 3 & $-32.0$ [$-47.0$, $-16.4$] & $-3.2$ [$-7.1$, $+0.7$] & $-0.12$ [$-0.17$, $-0.07$] \\
W1 (unseen robots) & 1 & $-26.0$ [$-38.3$, $-13.7$] & $-2.3$ [$-5.7$, $+1.0$] & $+0.01$ [$-0.04$, $+0.05$] \\
 & 2 & $-23.4$ [$-36.0$, $-10.9$] & $-2.8$ [$-5.9$, $+0.2$] & $+0.05$ [$+0.01$, $+0.10$] \\
 & 3 & $-27.3$ [$-40.4$, $-13.6$] & $-1.9$ [$-5.5$, $+1.5$] & $+0.08$ [$+0.03$, $+0.12$] \\
W3 (unseen combination) & 1 & $-31.6$ [$-45.8$, $-17.8$] & $-4.3$ [$-10.0$, $+1.1$] & $-0.04$ [$-0.09$, $+0.02$] \\
 & 2 & $-50.4$ [$-65.9$, $-35.5$] & $-7.6$ [$-12.9$, $-2.8$] & $+0.05$ [$-0.01$, $+0.10$] \\
 & 3 & $-46.2$ [$-63.8$, $-28.6$] & $-9.0$ [$-14.5$, $-3.8$] & $-0.02$ [$-0.07$, $+0.03$] \\
W4 ($N{=}32$) & 1 & $-167.3$ [$-218.6$, $-119.4$] & $-24.2$ [$-34.7$, $-14.5$] & $-0.15$ [$-0.18$, $-0.12$] \\
 & 2 & $-92.0$ [$-139.7$, $-43.3$] & $-7.8$ [$-16.1$, $+0.6$] & $-0.05$ [$-0.08$, $-0.02$] \\
 & 3 & $-49.2$ [$-97.7$, $-1.8$] & $+7.2$ [$-4.8$, $+19.1$] & $-0.13$ [$-0.16$, $-0.10$] \\
\bottomrule
\end{tabular}
}
\end{table}

\section{Training details}\label{app:train}

All values below are read from the training configuration of the reported runs.
MAPPO runs \res{cfg/iters} iterations of \res{cfg/eps_per_iter} episodes with Adam (learning rate \res{cfg/lr}), discount \res{cfg/gamma}, GAE parameter \res{cfg/lam}, clip ratio \res{cfg/clip}, \res{cfg/epochs} epochs per iteration, minibatches of \res{cfg/minibatch} agent-steps and entropy coefficient \res{cfg/ent}.
The reward weights are \res{rw/progress} for progress, \res{rw/arrive} for arrival, \res{rw/interv} for intervention, \res{rw/infeas} for infeasible QPs, \res{rw/viol} for violations and \res{rw/speed} for overspeed.
Training episodes use the W0 layout (design robot classes, \cls{Payload} and \cls{HazardousPayload}) with random positions along the lanes and random initial carrying states, \res{cfg/T} steps each, and evaluation episodes have 800 steps.
The arrival bonus is paid when a goal is reached and switched, so waiting at a pickup deferred by the activation check earns nothing.
Validation uses W0 episodes with seeds disjoint from the evaluation seeds.
The own features of an agent are its goal offset, its velocity, its payload offset, the keep-out its payload currently requires and its class embedding.
Each entry for another entity holds the relative position and velocity of the entity, the distance, the keep-out $D$, the margin between them and the class embedding of the entity, and an agent that carries a payload also sees chain-related entities measured from its payload.
The encoder uses attention pooling with 4 heads of total width 64 and a null token, and entry weights fade linearly to zero over the last 1\,m before $\rho$.
Class embeddings have dimension 8 and are tied along the taxonomy ($e_C=\sum_{C'\sqsupseteq C}\delta_{C'}$ with $\delta_C=0$ for classes absent from training), and the actor and critic heads are two-layer tanh MLPs of width 256.
Three training seeds are run, so the between-run part of the intervals in Section~\ref{sec:experiments} is coarse, and Table~\ref{tab:runs} gives each run separately.
The PD task controller commands $u_i=2(v_i^{\mathrm{des}}-v_i)$, with $v_i^{\mathrm{des}}$ pointing at the goal at speed $\min(\bar v_{\kappa(i)},\text{distance to the goal})$, and humans follow waypoints with the same law and $\|\dot v_h\|\le\bar a_h$.
Two reward configurations were trained, the weights above and a larger arrival bonus of 3 with intervention weight 0.003.
A rule fixed beforehand admits the configurations whose violating steps on W0 validation episodes are at most those of PD and takes the one with the most deliveries, which selected the configuration above.
The residual scale $s_i$ and its cap, used in all results reported here, were introduced in an earlier version of these experiments with other scenarios, in which the corrections of a policy trained on \cls{Drone}, scaled up by the larger $\bar a$ of \cls{LightDrone}, cost goals.

\end{document}